%% file: neurips_2026.tex
\documentclass{article}

 \usepackage[preprint]{neurips_2026}

\usepackage[utf8]{inputenc} 
\usepackage[T1]{fontenc}    
\usepackage{hyperref}       
\usepackage{url}            
\usepackage{booktabs}       
\usepackage{amsfonts}       
\usepackage{nicefrac}       
\usepackage{microtype}      
\usepackage{xcolor}         

\input{cmds}

\title{Direct Reasoning Optimization: Token-Level Reasoning Reflectivity Meets Rubric Gates for Unverifiable Tasks}

\author{%
  \normalfont
  \textbf{Yifei Xu}\textsuperscript{1}\thanks{Equal contribution.}\,\,\,
  \textbf{Tusher Chakraborty}\textsuperscript{1}\footnotemark[1]\,\,\,
  \textbf{Srinagesh Sharma}\textsuperscript{1}\,
  \textbf{Leonardo Nunes}\textsuperscript{1}\,
  \textbf{Swati Sharma}\textsuperscript{1}\, \\
  \textbf{Kate Drakos Demopulos}\textsuperscript{1}\,
  \textbf{Emre K{\i}c{\i}man}\textsuperscript{1}\,
  \textbf{Songwu Lu}\textsuperscript{2}\,
  \textbf{Ranveer Chandra}\textsuperscript{1} \\[4pt]
  \textsuperscript{1}Microsoft \quad
  \textsuperscript{2}University of California, Los Angeles \\
  \texttt{\{yifexu, tusher.chakraborty\}@microsoft.com}
}

\begin{document}

\maketitle

\input{0-abstract}
\input{1-intro}

\input{2-related}
\input{3-method}
\input{4-exp}

\input{5-conclusion}


\bibliographystyle{plainnat}
\bibliography{references}

\appendix
\newpage
\input{6-appendix}


\end{document}

%% file: cmds.tex
\usepackage{enumitem}
\usepackage{booktabs}
\usepackage{multirow}
\usepackage{graphicx}
\usepackage{subcaption}
\usepackage{makecell}
\usepackage{graphicx}
\usepackage{subcaption}
\usepackage{amsmath}
\usepackage{amssymb}
\usepackage{mathtools}
\usepackage{amsthm}
\usepackage[most]{tcolorbox}
\usepackage[capitalize,noabbrev]{cleveref}
\usepackage[textsize=tiny]{todonotes}

\theoremstyle{plain}

\theoremstyle{definition}

\theoremstyle{remark}

\newcommand{\myname}[1]{$\bf{DRO}$#1}
\newcommand{\rwdname}[1]{$\bf{R3}$#1}

\newcommand{\tusher}[1]{\textcolor{red}{}}

\newcommand{\bbb}[1]{

\noindent\textbf{#1}}
\newcommand{\yifei}[1]{\textcolor{orange}{}}

\newcommand{\para}[1]{ParaRev}
\newcommand{\rar}[1]{RaR-Medicine}
\newcommand{\gov}[1]{GovReport}
\newcommand{\nda}[1]{ContractNLI}
\newcommand{\finqa}[1]{FinQA}
\newcommand{\qmsum}[1]{QMSum}

%% file: 0-abstract.tex
\begin{abstract}
Reinforcement learning (RL) training of large language models (LLMs) on unverifiable tasks is challenging even when a reasonable-quality reference answer is available. We propose a constrained RL training framework that (i) optimizes a token-level dense \emph{Reasoning Reflection Reward} (\rwdname{}) aligned with reasoning quality, and (ii) enforces rubric-gating as feasibility constraints at the rollout group level. \rwdname{} measures the model's token-level certainty of a reference answer under its chain-of-thought (CoT) prefix, and selectively emphasizes tokens with high cross-rollout variance, which we call \emph{reasoning-reflective tokens}, that would otherwise be diluted by the bulk of low-variance tokens. The same variance signal also drives a filter that discards queries with insufficient signal for comparative learning. Rubric-gating complements \rwdname{} by operationalizing principled task criteria as hard accept/reject checks on final answers. Empirically, across four datasets spanning scientific writing, medicine, legal contracts, and finance, our framework outperforms strong baselines, achieves faster, more sample-efficient learning, and respects feasibility constraints.

\end{abstract}

%% file: 1-intro.tex
\section{Introduction}

Large Language Models (LLMs) have recently demonstrated strong reasoning capabilities, particularly in mathematics, programming, and scientific problem solving~\citep{guo2025deepseek, jaech2024openai, zeng2024scaling, yang2025qwen3}. These advances are often achieved through Reinforcement Learning with Verifiable Rewards (RLVR), where reward signals, such as matching reference answers or passing unit tests, provide stable supervision during policy optimization~\citep{schulman2017proximal, liu2025understanding, yu2025dapo}. Motivated by the success of RLVR in structured domains, there is growing interest in applying similar techniques to open-ended, long-form tasks such as scientific document revision~\citep{jourdan2025pararev}, medical question answering~\citep{gunjal2025rubrics}, creative writing~\citep{lu2025writing}, and analytical summarization~\citep{huang2021efficient}, among others. For a significant subset of these tasks, a high-quality reference answer is available, yet outcomes still cannot be programmatically verified, making reward design on such unverifiable tasks substantially more challenging~\citep{zhao2025absolute, su2025crossing, gunjal2025rubrics}.

We propose a \emph{constrained RL}~\citep{ray2019benchmarking} training design that (i) optimizes a token-level dense reward aligned with reasoning quality, and (ii) enforces rubric-gating as principled feasibility constraints at the rollout group level (a group = $k$ rollouts per query). To this end, we introduce the Reasoning Reflection Reward (\rwdname{}), a token-level dense reward tailored for unverifiable tasks with long-form answers spanning hundreds of tokens (and that remains effective on short-form ones). Since chain-of-thought (CoT) reasoning acts as a latent prefix that conditions the final answer, we measure the model's token-level certainty of a reference answer under this prefix, capturing how likely the generated CoT reasoning is to yield the desired answer. We observe that the most informative tokens in reference answers, often not lexically distinctive, are those whose self-certainty varies substantially with the CoT, whereas the majority is largely unaffected by reasoning. We refer to these tokens as \emph{reasoning-reflective tokens}. In a uniform average of self-certainties, however, their contribution is diluted by the low-variance majority. \rwdname{} therefore identifies and up-weights these reasoning-reflective tokens leveraging their \emph{cross-rollout variance}, producing a more focused reward signal that sharpens the contrast across rollouts in a group. The same cross-rollout variance is, in turn, a query-level signal: a group whose tokens all collapse to near-zero variance carries no comparative information for policy improvement, motivating a variance-driven query filter that further stabilizes training. Yet token-level dense rewards alone remain vulnerable to reward hacking: without lexical or semantic supervision, an actor model may produce a rollout group whose answers are uniformly low-quality yet still exhibit relative differences under the token-level dense metric, misleading the gradient update.

\begin{figure*}[t]
  \centering
  \includegraphics[width=0.95\textwidth]{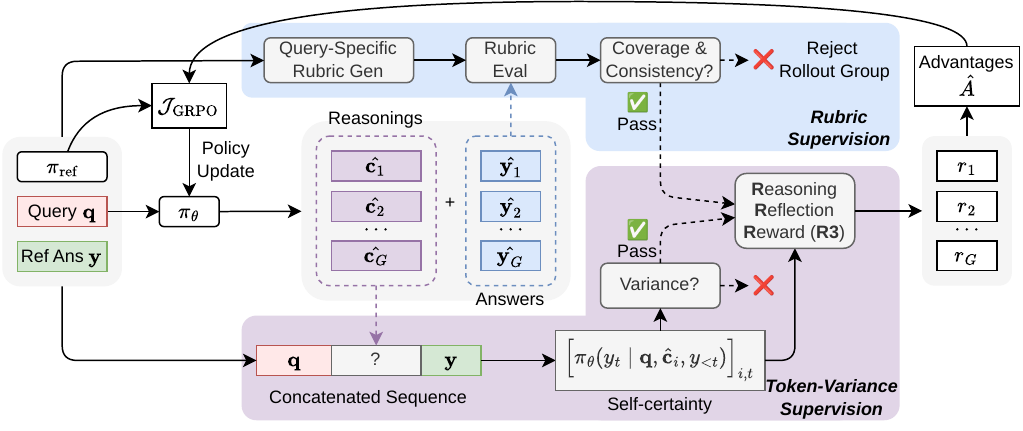}
  \caption{\textbf{\myname{} as a constrained RL setup: token-level dense reward with rubric gating.} \rwdname{} provides a dense, token-level reward for CoT traces by emphasizing reasoning-reflective tokens identified through cross-rollout variance; rubric supervision enforces \emph{feasibility} lexical and semantic constraints on final answers through rollout group-level rejection; and the same cross-rollout variance signal drives a query filter that skips groups with insufficient signal for comparative learning.}
  \label{fig:overview}
  \vspace{-0.1in}
\end{figure*}

Rubrics provide precisely this supervision. Yet directly converting rubric judgments into dense rewards is brittle: crafting fine-grained criteria at scale is labor-intensive~\citep{liu2025openrubrics, he2025advancedif}, LLM-generated rubrics can be inconsistent when atomic or evolving criteria are needed~\citep{xie2025auto, saad2025shrinking}, and multi-criterion aggregation compounds noise from per-criterion scores and LLM-assigned weights~\citep{zhou2025breaking, huang2025reinforcement}. Attempts to stabilize scoring via per-criterion evaluation and majority voting substantially increase compute cost~\citep{wu2025rlac}. Conversely, foregoing fine-grained rubrics leads to score saturation and over-regularization to superficial features~\citep{shao2025dr, rezaei2025online, zhou2025breaking, huang2025reinforcement}. Nevertheless, \emph{rubrics reliably encode principled guardrails}~\citep{bai2022constitutional}. We therefore use rubrics for \emph{gating}: a rollout group is accepted or rejected based on essential task criteria, rather than converted into dense rewards. While \rwdname{} primarily targets the quality of CoT reasoning, rubric-gating provides complementary supervision that ensures final answers satisfy fundamental task constraints.

\noindent\textbf{Summary of Contributions:} In this paper, we make the following contributions.
\begin{itemize}[leftmargin=*,topsep=2pt,itemsep=3pt,parsep=3pt]
    \item We present a \emph{constrained RL} training framework built on Group Relative Policy Optimization (GRPO)~\citep{shao2024deepseekmath} for unverifiable tasks that employs the token-level dense reward with rubric-gated constraints. Both rubric-gating decisions and reward computations are derived from the same reference policy used during RL training, eliminating any reliance on external judges or reward models. We refer to this framework as \emph{Direct Reasoning Optimization} (\myname{}) (Fig.~\ref{fig:overview}).
    \item We design \rwdname{}, a token-level reward that highlights reasoning-reflective tokens via their \emph{cross-rollout variance}, and show that this same variance signal naturally extends to a query-level filter that further stabilizes training and improves sample efficiency. We provide analytical and empirical evidence for both.
    \item We evaluate \myname{} on four datasets spanning distinct domains, three of which require long-form answers (hundreds of tokens). \myname{} consistently outperforms baselines on downstream metrics (Table~\ref{tab:main_results}), reaches comparable performance up to $2$--$3\times$ faster (Table~\ref{tab:r3_faster_learning}), satisfies rubric-based feasibility constraints (Fig.~\ref{fig:rubric_convergence}), exhibits cross-dataset knowledge transfer (Appendix~\ref{app:healthbench}), and yields more stable, sample-efficient training (Section~\ref{sec:eval_r3_vs_plain}, \ref{sec:eval_variance_based_filtering}).
\end{itemize}

%% file: 2-related.tex
\vspace{-2mm}

\section{Related Work}
\subsection{Reinforcement Learning with Verifiable Rewards}
Reinforcement Learning from Verifiable Rewards (RLVR) improves LLM performance in domains such as coding and mathematics, where outcomes can be unambiguously verified, eliminating the need for learned reward models~\citep{lambert2024t, liu2025understanding, hu2025open, luo3deepcoder, liu2025code}.
To optimize policies under RLVR, \citet{shao2024deepseekmath} proposed Group Relative Policy Optimization (GRPO), a PPO~\citep{schulman2017proximal} variant that computes advantages by comparing a group of $G$ sampled outputs $\{\textbf{o}_i\}_{i=1}^G$ for the same prompt with rewards $\{r_i\}_{i=1}^G$:
\begin{equation}
\label{eq:grpo_advantage}
 \hat A_{i,t} = \frac{r_i - mean(\left\{r_i\right\}_{i=1}^G)}{std(\left\{r_i\right\}_{i=1}^G)}    
\end{equation}
where $r_i$ is typically a verifiable outcome reward (e.g., 1 for a correct answer, 0 otherwise). We defer the full GRPO objective to Appendix~\ref{app:grpo_objective} and use GRPO as our optimization backbone. Another RLVR work, the \emph{80/20 rule}~\citep{wang2025beyond}, exploits a small subset of high-entropy CoT tokens for gradient updates; we discuss the contrast in Section~\ref{sec:r3_design}.

\subsection{Reinforcement Learning on Unverifiable Tasks}
Over the past year, considerable efforts have been made to extend RLVR to unverifiable, open-ended reasoning tasks. One line of work trains general-purpose reward models to supervise reasoning optimization~\citep{chen2025rm, liu2025inference, su2025crossing}, introducing the overhead of an additional reward model during RL training.

\noindent\textbf{Rewards from self-certainty.}
A complementary line uses internal model feedback such as self-certainty as the reward signal, eliminating external verifiers. Several concurrent studies~\citep{zhao2025absolute, xu2025genius, zhao2025learning, zuo2025ttrl} rely exclusively on intrinsic feedback without reference answers, while reference-based variants incorporate reference outcomes to estimate reasoning quality~\citep{chen2024language, tang2025learning}. Among them, VeriFree~\citep{zhou2025reinforcing} and RLPR~\citep{yu2025rlpr} weight each reference token uniformly (using average log-probability and average probability, respectively), while \citet{wang2026native} weights tokens by their probability under a frozen baseline. \rwdname{} instead weights tokens by their \emph{cross-rollout variance}, directly capturing how much the CoT shifts that token's certainty (Section~\ref{sec:r3}). Beyond per-token rewards, none of these methods address reward hacking through lexical or semantic supervision, which we tackle via rubric-gating (Section~\ref{sec:rubric_gating}). Moreover, existing self-certainty methods are evaluated only on tasks with short-form answers or programmatic verifiers (e.g., multiple choice, verifiable instructions), leaving long-form, unverifiable tasks untested.

\noindent\textbf{Rewards from rubrics.}
Rubrics provide structured evaluation criteria for model outputs~\citep{pathak2025rubric}. While human-authored rubrics can be reliable, writing them at scale is labor-intensive~\citep{liu2025openrubrics, he2025advancedif, xie2025auto} and inherently incomplete, as new failure modes emerge during training~\citep{shao2025dr, rezaei2025online}. Recent work uses LLMs to generate rubrics and assign weights~\citep{viswanathan2025checklists}, but both remain unreliable. Existing rubric-based evaluation follows two regimes: query-agnostic (offline) rubrics, e.g., RaR~\citep{gunjal2025rubrics}, use static task-level criteria that are easy to maintain but prone to score saturation and over-regularization, inducing mode collapse during RL~\citep{huang2025reinforcement}; query-specific (online) rubrics, e.g., RLER~\citep{shao2025dr}, tailor criteria per query (generated directly from the task or contrastively across candidates) and provide higher-resolution feedback but risk drifting from the original task definition or amplifying idiosyncrasies of sampled outputs. Rewards are then obtained by prompting an LLM judge to score each criterion: scores are inconsistent when criteria are bundled, and computationally expensive~\citep{wu2025rlac} and weight-sensitive when scored separately.

%% file: 3-method.tex
\section{Reasoning Reflection Reward (\rwdname{})}
\label{sec:r3}

\begin{figure*}[t]
  \centering
  \includegraphics[width=0.99\textwidth]{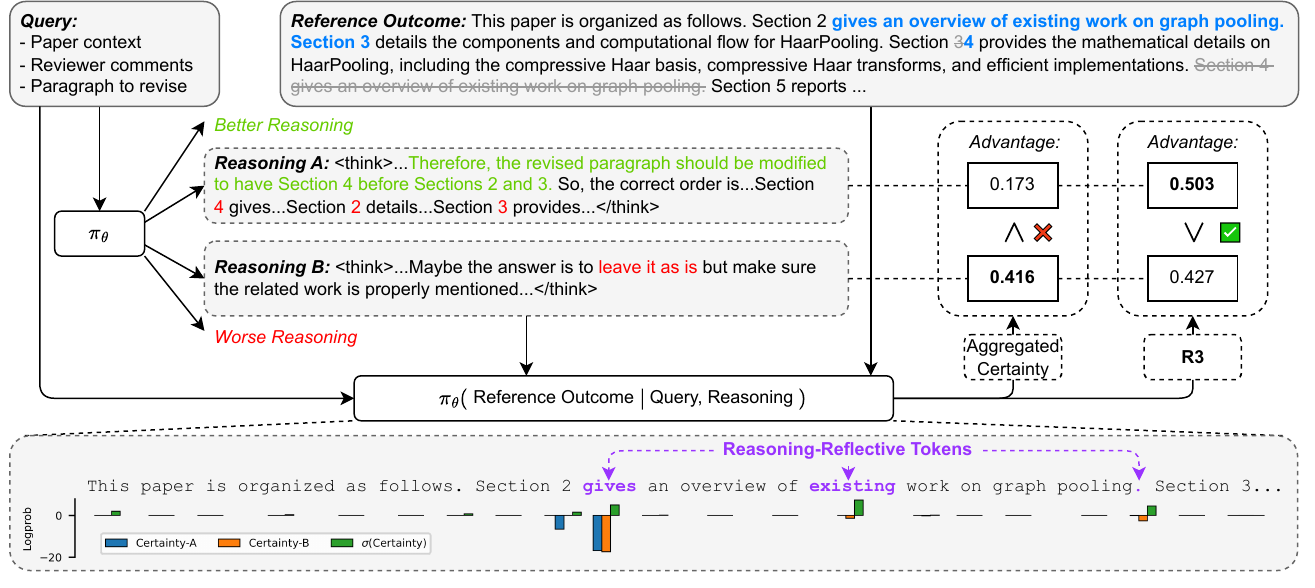}
  \caption{Illustrative example of Reasoning Reflection Reward (\rwdname{}). For the paper revision task, the model is prompted to revise a paragraph based on reviewer comments (upper left). \rwdname{} computes per-token self-certainty (log-probabilities) in the reference revision (upper right) for each sampled reasoning trace, and identifies reasoning-reflective tokens by their cross-rollout standard deviation $\sigma$.
  In this example, Reasoning A correctly identifies that Section 4 (overview) has been moved earlier and adjusts the paragraph structure accordingly, with a minor omission of section numbers. Reasoning B gives up. While a vanilla aggregate of certainty prefers B over A due to A's lower certainty on the token ``\texttt{2}'', \rwdname{} successfully aligns with the desired ranking by up-weighting high-$\sigma$ tokens such as ``\texttt{gives}'', ``\texttt{existing}'' and ``\texttt{.}'' that better reflect reasoning effectiveness.}
  \label{fig:reward_explained}
  \vspace{-0.1in}
\end{figure*}

We construct a token-level dense reward for optimizing an LLM's chain-of-thought (CoT) reasoning in RL on unverifiable tasks with long-form (100s of tokens) reference answers.

\subsection{Self-certainty and Reasoning-Reflective Tokens}

A reasoning LLM's output typically consists of a CoT trace followed by a final answer. Because generation is autoregressive, the CoT acts as a latent prefix conditioning the answer distribution~\citep{chen2025reasoning, chen2024language}. Formally, with query $\mathbf{q}$ and generated CoT $\hat{\mathbf{c}}$, the answer is sampled from $\pi(\cdot \mid \mathbf{q}, \hat{\mathbf{c}})$. Intuitively, well-structured reasoning increases the likelihood of a high-quality answer, while flawed reasoning reduces it. This suggests a natural dense reward: reward CoT traces by how likely they are to produce the reference outcome. We operationalize this via \emph{self-certainty}~\citep{gupta2024language, kauf2024log}, the likelihood $\pi(\mathbf{y} \mid \mathbf{q}, \hat{\mathbf{c}})$ the model assigns to the reference $\mathbf{y}$ under its CoT prefix; this scalar can in principle serve as $r$ in Eq.~\ref{eq:grpo_advantage} (we show it is not equivalent to SFT in Appendix~\ref{appendix:grpo_self_cert_not_sft}). However, long-form answers contain many tokens with individual likelihoods, raising a central design question: \emph{how should we aggregate token-level self-certainty into a single reward $r$ that meaningfully reflects reasoning quality?}

\subsubsection{Aggregation of Self-certainties}
\label{sec:aggregation}
A model's self-certainty over reference token $y_j$ conditioned on CoT $\hat{\mathbf{c}}_i$ is the conditional probability $\pi(y_j \mid \mathbf{q}, \hat{\mathbf{c}}_i, \mathbf{y}_{<j})$. A natural reward aggregates these across the reference sequence, e.g., $r = \sum_{j=1}^{\lvert \mathbf{y} \rvert} \log \pi(y_j \mid \mathbf{q}, \hat{\mathbf{c}}_i, \mathbf{y}_{<j})$ (or its raw-probability variant). To assess whether such plain aggregation discriminates CoT traces, we conduct a case study on ParaRev, where the task is to revise a paragraph in response to reviewer comments, not all of which are necessarily relevant (Section~\ref{sec:datasets}). For one query, we sample $16$ outputs and manually rank them by how well their CoTs and final answers address the relevant comments and align with the reference. Fig.~\ref{fig:reward_explained} presents two representative traces with advantages from plain aggregate self-certainty. Notably, the resulting advantages correlate only weakly with quality, and in this instance even rank a lower-quality CoT trace above a higher-quality one.

To understand this behavior, we inspect the token-level distributions in Fig.~\ref{fig:reward_explained}: most reference tokens receive similar likelihoods regardless of the preceding CoT, while only a small subset (three tokens here) exhibits appreciable variance with the reasoning trace. We call these \emph{reasoning-reflective tokens}. However, sequence-wide aggregation dilutes their influence, masking informative differences between high- and low-quality CoT traces; this becomes especially pronounced when reasoning-reflective tokens are few relative to total length. \citet{wang2025beyond} similarly find that only a small fraction of CoT tokens are high-entropy and serve as critical branching points. Suppression of such informative token-level signals through sequence-wide aggregation has also been observed in model cascading and hallucination detection~\citep{chen2025enhancing}.

\subsubsection{Analytical View: Dilution Under Groupwise $z$-Scoring}
GRPO computes advantages via groupwise $z$-scoring (Eq.~\ref{eq:grpo_advantage}), so we want $z$-scores to track variations in reasoning-reflective tokens rather than be overwhelmed by uninformative ones. Our analysis in Appendix~\ref{appendix:math_analysis_r3} shows that with sequence-wide averaging over $t$ reference tokens, the contribution of a small informative subset shrinks rapidly: $z$-scores decay as $1/\sqrt{t}$ under independent noise and as $1/t$ under positive correlation. This \emph{dilution effect} explains why plain self-certainty struggles to rank reasoning traces in long-form generation.

\begin{figure}[t]
  \centering
  \includegraphics[width=0.45\textwidth]{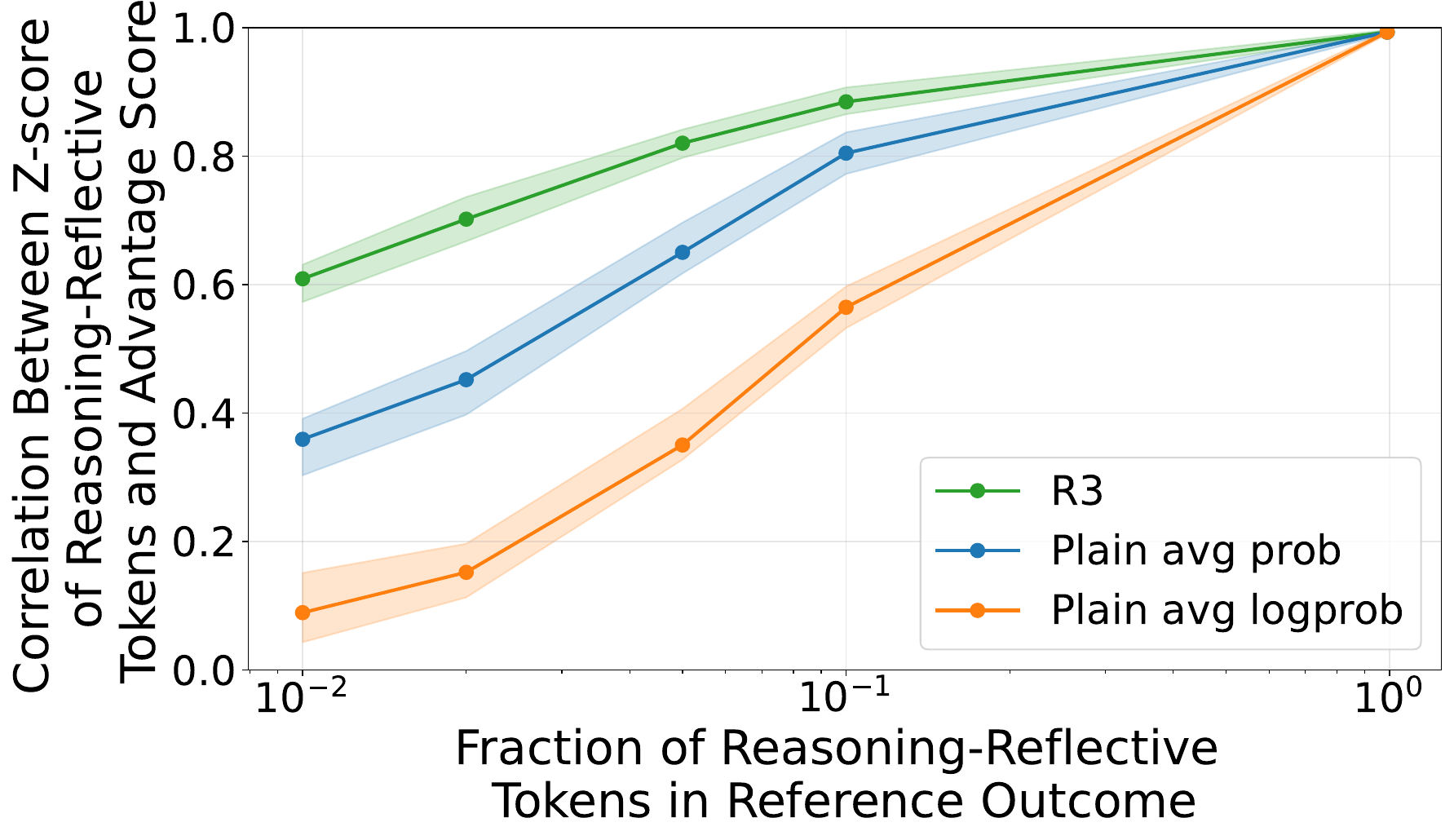}
  \caption{\textbf{Simulation results} comparing correlation between rollout advantages and reasoning-reflective token signals under different aggregation methods. Plain average log-probability exhibits severe degradation due to noisy low-probability tokens, average probability performs moderately better but still suffers dilution, while \rwdname{} consistently maintains higher correlation across sequence lengths and reflective-token fractions.}
  \label{fig:dilution_effect_scoring_comparison}
  \vspace{-0.1in}
\end{figure}

We corroborate this with a controlled simulation (Appendix~\ref{appendix:sim_setup}) calibrated from manual inspection of all evaluation datasets. Per rollout group, it measures the correlation between (i) advantages under different aggregations and (ii) $z$-scores of reasoning-reflective tokens. Consistent with our inspection, the fraction of reasoning-reflective tokens is typically very small ($\approx 2$--$3\%$, and frequently $<\!1\%$ for corpora such as \nda{}~\citep{koreeda2021contractnli}). Moreover, since our scoring $\pi(y_j \mid \mathbf{q}, \hat{\mathbf{c}}_i, \mathbf{y}_{<j})$ is autoregressive, earlier reasoning-reflective tokens influence the likelihoods of later ones; this dependence reduces the number of independently informative tokens, further exacerbating dilution. As Fig.~\ref{fig:dilution_effect_scoring_comparison} shows, plain average log-probability correlates worst with reasoning-reflective $z$-scores, primarily because extremely low-probability ($<\!0.01$) tokens (typically from style/formatting misalignment) induce large noisy variance on the log scale (e.g., $-4$ to $-25$) from negligible probability changes. Aggregating on the probability scale alleviates this and yields higher correlations. Nevertheless, even in probability space, sequence-wide averaging still dilutes when reasoning-reflective tokens are heavily outnumbered, as in long-form references dominated by lexical synonyms, function words, structural tokens, and globally predictable factual units.

\subsection{\rwdname{} Emphasizes Reasoning-Reflective Tokens}
\label{sec:r3_design}
To mitigate dilution, we introduce the Reasoning Reflection Reward, which selectively emphasizes reasoning-reflective tokens. As Fig.~\ref{fig:reward_explained} shows, many such tokens are not semantically salient in isolation and elude lexical heuristics. We instead identify them comparatively: reasoning-reflective tokens are those whose likelihoods exhibit high variability across CoT prefixes, directly capturing upstream reasoning influence. This comparative spirit also underlies GRPO's connection to preference-based reward modeling~\citep{shao2024deepseekmath}. \rwdname{} weights each reference token's self-certainty by its variability, with the reward for rollout $i$ given by:
\begin{equation}
\label{eq:r3}
 r_i^{\text{R3}} = \sum_{j=1}^{|\mathbf{y}|}
\underbrace{\frac{e^{\omega \sigma_j}}
{\sum_{k=1}^{|\mathbf{y}|} e^{\omega \sigma_k}}}_{\text{softmax weight}}
\cdot 
\operatorname{clip}\!\left(
\pi(y_j \mid \mathbf{q}, \hat{\mathbf{c}}_i, \mathbf{y}_{<j}),
\, \lambda_{\text{low}},\, \lambda_{\text{high}}
\right)
\end{equation}
where $\sigma_j$ is the standard deviation of token $j$'s self-certainty across sampled CoT traces, and $\omega$ is a temperature-like hyperparameter controlling sharpness. We use a softmax rather than direct normalization because it prevents overly sharp emphasis on reasoning-reflective tokens, allowing $\omega$ to modulate how strongly they are highlighted (see $\omega$ tuning in Section~\ref{sec:eval_r3_vs_plain}). We use probability rather than log-probability since, per the analysis above, it yields more stable correlations. Clipping to $[\lambda_{\text{low}}, \lambda_{\text{high}}]$ further mitigates dilution: tokens with consistently high probabilities (e.g., $\gtrsim 0.85$) are already reliably predicted and their small fluctuations would only downweight informative mid-range tokens (Appendix~\ref{appendix:math_analysis_r3}); tokens with consistently low probabilities ($\lesssim 0.05$) typically reflect lexical/stylistic mismatches and contribute noise. Manual inspection across datasets confirms that a large portion of uninformative tokens fall into these extreme regions.

Looking back at Fig.~\ref{fig:dilution_effect_scoring_comparison}, \rwdname{} achieves substantially higher correlation with reasoning-reflective tokens than both average log-probability and probability, particularly when reasoning-reflective tokens constitute $<\!10\%$ of the reference. Note that, for cleaner curves, the simulation also imposes positive correlation among reasoning-reflective tokens; relaxing this would only widen \rwdname{}'s margin. This selective amplification also yields faster learning than plain aggregation by producing higher-contrast advantages, raising the gradient signal-to-noise ratio (Section~\ref{sec:eval_r3_vs_plain}). The \emph{80/20 rule}~\citep{wang2025beyond} reports a parallel finding in RLVR, but on a different axis: it selects \emph{CoT tokens} (by entropy) for \emph{gradient masking}, whereas \rwdname{} selects \emph{reference-outcome tokens} (by cross-rollout variance) for \emph{reward shaping}. The two are therefore orthogonal and complementary, and we apply the 80/20 rule on top of \rwdname{} in our experiments.

\section{Direct Reasoning Optimization with \rwdname{} and Rubric-Gating}
\label{sec:DRO}
In addition to \rwdname{}, \myname{} (Fig.~\ref{fig:overview}) introduces two complementary components: \emph{rubric-gating}, which enforces essential task-specific \emph{feasibility constraints} on final answers, and \emph{variance-based filtering}, which prevents spurious groupwise $z$-score amplification when reward variance is low.

\subsection{Enforcing Rubric Supervision}
\label{sec:rubric_gating}
\rwdname{} is discriminative within a rollout group, but a critical limitation remains: even when \emph{all} answers in a group violate core task principles, a token-level reward still produces well-structured advantages and misguides the policy gradient. Groupwise standardization (Eq.~\ref{eq:grpo_advantage}) further amplifies these spurious differences among uniformly low-quality rollouts. On ParaRev, for instance, we observe rollout groups whose responses perform no meaningful revision \emph{at all}, yet still receive distinguishable advantages, eventually triggering model collapse (Appendix~\ref{app:no_revision}). These observations motivate a \emph{constrained RL} design to enforce feasibility constraints on \emph{final answers}. While rubric-based \emph{dense} scoring is unstable, rubrics remain reliable for enforcing task principles in lexical and semantic spaces~\cite{bai2022constitutional}.

We enforce these constraints via rubric-gating, which implements rejection of rollout groups (note that the query is not dropped from the training set) based on essential query-specific rubrics. For each query, the reference-policy LLM proposes a compact set of yes/no rubrics, and we verify that the reference answer satisfies each one to avoid ill-posed constraints (prompts in Appendix~\ref{appendix:rubric_gen_prompt_dro}). We then apply two gates to a rollout group: (i) \textbf{Coverage gate:} every rubric must be satisfied by at least $\mu$ rollouts in the group, ensuring feasibility compliance with a tunable strictness via $\mu$; (ii) \textbf{Consistency gate:} the top $\rho\%$ of rollouts ranked by $r_i^{\text{R3}}$ must each satisfy at least $\nu\%$ of the rubrics. This tests whether the rollouts most responsible for driving policy updates are also constrained at a minimum coverage level $\nu$, thereby preventing high \rwdname{} scores from arising in outputs that violate essential constraints. This separates feasibility determination from reward ranking and aligns the dense reward's update carriers with rubric validity. Together, $(\rho,\nu)$ balance robustness and exploration, mitigating reward hacking without over-regularizing to LLM-generated principles.

\subsection{Variance-based Filtering for Learning Stability}
\label{sec:variance_filtering}
In GRPO, stable optimization requires sufficient \emph{variance} in reward signals within each rollout group~\cite{liu2025understanding}. However, with dense rewards and groupwise standardization, even small absolute differences can be spuriously amplified in the advantages when the underlying variance is low, destabilizing gradient updates. To mitigate this, we apply a \emph{variance-based filter} that scores every query in the corpus once under the reference policy and retains only those whose rollout groups carry strong comparative signal. For a query's rollout group $G$, let $\sigma_{G,j}$ be the cross-rollout standard deviation of token $j$'s self-certainty and $J_{\text{top}}$ the top $10\%$ of tokens by $\sigma_{G,j}$. We score $G$ by its \emph{high-variance score} $\bar{\sigma}_G = \frac{1}{|J_{\text{top}}|}\sum_{j \in J_{\text{top}}} \sigma_{G,j}$, rank queries by $\bar{\sigma}_G$, and keep the top fraction ($q\!\approx\!10\%$) plus a small additional fraction to preserve hard cases. Empirically, we find that \emph{quality outweighs quantity}: a few hundred high-variance groups yield better learning, whereas training on whole unfiltered corpora precipitates model collapse.

%% file: 4-exp.tex
\vspace{-1.5mm}
\section{Experiments}
\vspace{-1.5mm}







    

\vspace{-1.5mm}
\subsection{Experimental Setup}
\vspace{-1.5mm}

\bbb{Datasets.}
We evaluate \myname{} on four datasets spanning diverse domains and task formats, with an emphasis on long-form generation: \textbf{\para{}} (scientific paragraph revision) and \textbf{\rar{}} (medical QA) are open-ended with reference answers; \textbf{\nda{}} (legal contract review) is less open-ended as quality hinges on label correctness; and \textbf{\finqa{}} (financial numerical reasoning) is verifiable.

\bbb{Baselines.} To isolate the contribution of the reward design, we adopt four baseline categories: 
\textbf{(1) Base:} off-the-shelf models without RL; 
\textbf{(2) RL-F1:} ROUGE-L F1 reward as a standard lexical-similarity objective (for \finqa{}, a verifiable task, we instead use answer correctness as the reward, serving as an upper bound); 
\textbf{(3) Avg Prob (RLPR) / Logprob (VeriFree):} plain probability or log-probability aggregation over outcome tokens (Section~\ref{sec:aggregation}), as used in RLPR~\citep{yu2025rlpr} and VeriFree~\citep{zhou2025reinforcing}; 
\textbf{(4) Rubric reward (RLER):} query-specific rubrics combined with discriminative evolving rubrics from concurrent work RLER~\citep{shao2025dr} (Appendix~\ref{appendix:rubric_baseline}).
We use GRPO as the default RL method for all baselines so that differences arise from reward design rather than other components. RLPR's standard-deviation filter is compared separately in Appendix~\ref{app:rlpr_filter_comparison}, and VeriFree's RLOO-style estimator is replaced with GRPO for the same reason.

Details on datasets, preprocessing, metrics, training/inference configurations, \myname{}-specific hyperparameters, and training/inference efficiency are provided in Appendix~\ref{app:eval_setup}.



\vspace{-1.5mm}
\subsection{Main Results}
\vspace{-1.5mm}

Table~\ref{tab:main_results} reports \myname{} against baselines and variants on each dataset. We summarize the main findings here; detailed ablations follow.

\bbb{\myname{} consistently outperforms RL baselines.} Across all four datasets, \myname{} attains the best performance with clear margins over every RL baseline (rows \#2--\#5). On \para{} and \rar{}, RL-F1 and Avg Logprob fail to even improve the base. On \finqa{}, \myname{} approaches RL with an oracle correctness reward (row \#2), indicating effectiveness even on short-form, verifiable tasks where dense self-certainty is typically considered redundant. ROUGE-L F1 against the reference is in Appendix~\ref{app:rouge_l} for completeness.

\bbb{\myname{} outperforms supervised and distillation baselines.} A natural question is whether the improvements arise from RL itself or could be recovered by simpler supervised/imitation-based recipes. We compare against two such alternatives on \para{} and \rar{}: SFT on the reference, and on-policy distillation from a much larger Qwen3-235B-A22B teacher into the same Qwen3-14B student. \myname{} substantially outperforms both, and SFT even \emph{degrades} on \para{} (Appendix~\ref{app:sft_distillation}).

\bbb{Combined supervision beats rubric or token-level reward alone.} The rubric-based reward (row \#5) and \rwdname{} (row \#6) each improve over the base, but rubric-gated \rwdname{} (row \#7) outperforms both: \rwdname{} discriminates within a rollout group while rubric-gating prevents updates from groups that uniformly violate task principles, jointly reducing reward hacking.

\bbb{Gains transfer to an independent medical benchmark.} To test whether \myname{} reshapes reasoning behavior beyond its training distribution, we evaluate \rar{}-trained checkpoints on the independently constructed HealthBench benchmark~\citep{arora2025healthbench} (Appendix~\ref{app:healthbench}). \myname{} yields gains of $+2.1$ for Qwen3-4B and $+0.6$ for Qwen3-14B (the latter already near GPT-4.1 level on HealthBench), indicating cross-benchmark transfer rather than dataset-specific memorization.

\bbb{\myname{} retains gains under an imperfect model-generated reference answers.} To evaluate \myname{}'s sensitivity to reference quality, we replace the human reference with a model-generated one (best-of-4 from Claude Sonnet~4.6) to simulate an imperfect, off-distribution reference, and re-train \myname{} with all other components unchanged (Appendix~\ref{app:imperfect_ref}). The model-reference variant still improves over the base on \para{} ($57.4$\%) and \rar{} ($55.1$\%), with smaller magnitudes than under the GT reference ($63.7$\% / $57.6$\%). Reference quality affects the magnitude, not the existence, of the improvement.

\begin{table*}[t]
  \centering
  \caption{\textbf{Main results.} Full \myname{} provides the best performance across datasets.}
  \label{tab:main_results}
  \setlength{\tabcolsep}{2pt} 
  \renewcommand{\arraystretch}{1.1}
  \small
  \begin{tabular}{c lcc cccc}
    \toprule
    & \multicolumn{3}{c}{\textbf{Method}} & \multicolumn{4}{c}{\textbf{Dataset}} \\
    \cmidrule(lr){2-4} \cmidrule(lr){5-8}

    \multirow{2}{*}{\textbf{\#}} &
    \multirow{2}{*}{\textbf{Reward}} &
    \multicolumn{2}{c}{\textbf{Rollout-Group Rejection}} &
    \multirow{2}{*}{\begin{tabular}[c]{@{}c@{}}\textbf{\para{}}\\[-1pt]\footnotesize WR vs. Base \end{tabular}} &
    \multirow{2}{*}{\begin{tabular}[c]{@{}c@{}}\textbf{\rar{}}\\[-1pt]\footnotesize WR vs. Base \end{tabular}} &
    \multirow{2}{*}{\begin{tabular}[c]{@{}c@{}}\textbf{\nda{}}\\[-1pt]\footnotesize Macro-F1\end{tabular}} &
    \multirow{2}{*}{\begin{tabular}[c]{@{}c@{}}\textbf{\finqa{}}\\[-1pt]\footnotesize Accuracy\end{tabular}} \\
    \cmidrule(lr){3-4}
    & & \textbf{Rubric} & \textbf{Variance} & & & & \\
    \midrule

    1 & Base        & --         & --         & 50.0 & 50.0  & 68.1 & 58.1 \\
    2 & RL-F1       & $\times$   & $\times$   & 40.8 & 46.7  & 73.4 & 69.0\textsuperscript{1} \\
    3 & Avg Logprob (VeriFree) & $\times$   & $\times$   & 49.5 & 48.0  & 73.6 & 66.4 \\
    4 & Avg Prob (RLPR)    & $\times$   & $\times$   & 54.3 & 52.3  & 78.2 & 66.8 \\
    5 & Rubric (RLER)     & $\times$   & $\times$   & 55.9 & 53.4  & 75.4 & --\textsuperscript{2} \\
    \midrule

    6 & \multirow{2}{*}{\rwdname{}} 
      & $\times$   & $\times$   & 57.8 & 54.2  & 80.6 & 67.1 \\
    7 &             
      & \checkmark & $\times$   & 61.6 & 56.1  & 80.8 & --\textsuperscript{2} \\
    \midrule

    8 & \textbf{Full \myname{}} 
      & \textbf{\checkmark} & \textbf{\checkmark} 
      & \textbf{63.7} & \textbf{57.6}  & \textbf{84.5} & \textbf{68.4} \\
    \bottomrule
  \end{tabular}

  \vspace{2pt}
  \footnotesize
  \begin{minipage}{\linewidth}
  \raggedright
  \textsuperscript{1}On \finqa{}, this score is obtained using an oracle correctness reward rather than RL-F1 and should be viewed as an upper bound.\\
  \textsuperscript{2}We do not apply rubric-based methods on \finqa{} because answers are short-form and verifiable.
  \end{minipage}
\end{table*}

\vspace{-1.5mm}
\subsection{Effect of \rwdname{}}
\label{sec:eval_r3_vs_plain}
\vspace{-1.5mm}

\bbb{Token-variance weighting helps most where dilution is most severe.}
\label{sec:eval:r3}
\rwdname{} consistently outperforms Avg Prob and Avg Logprob (Table~\ref{tab:main_results}, row \#6 vs.\ \#3, \#4). Gains are smaller on \nda{} (JSON-formatted, mostly stable high self-certainty) and \finqa{} (short answers, most tokens informative), where dilution is naturally limited. Consistent with Fig.~\ref{fig:dilution_effect_scoring_comparison}, Avg Prob outperforms Avg Logprob due to the latter's sensitivity to extremely low-probability tokens.

\bbb{\rwdname{} does not require a strict CoT--answer boundary.}
The main pipeline scores the reference conditioned on the sampled CoT alone, relying on a CoT/answer delimiter (e.g., \verb|<think>|). To test whether this boundary is essential, we ablate slicing by replacing the prefix with the \emph{full sampled completion} (CoT + sampled answer) and re-train under otherwise identical settings (Appendix~\ref{app:no_slicing}). The unsliced variant performs nearly identically on \para{} ($57.8$ vs.\ $57.5$) and \rar{} ($54.2$ vs.\ $54.3$), indicating that \rwdname{} measures how much the sampled prefix raises the reference-answer likelihood overall and applies to models without explicit CoT delimiters.

\bbb{Reasoning-reflective tokens deliver faster learning at matched per-step cost.}
With rubric-gating and variance-based filtering disabled for fairness, \rwdname{} reaches comparable downstream performance in $2$-$3{\times}$ fewer training steps than Avg Prob on \para{}, \rar{}, and \nda{}, with a smaller gap on \finqa{} (Appendix~\ref{app:r3_faster_learning}); per-step throughput and peak memory are comparable (Appendix~\ref{app:efficiency}), so step savings translate to wall-clock savings. Increasing $\omega$ accelerates early learning, but very high values ($\omega>4$) over-sharpen the softmax onto a tiny token subset, magnifying noise; we use $\omega\in[1,2]$ across datasets, with isolated $\omega$ and clipping ablations in Appendix~\ref{app:omega_clipping}.

\vspace{-1.5mm}
\subsection{Effect of Rubric Supervision}
\label{sec:eval_rubric_gating}
\vspace{-1.5mm}

\bbb{Rubric-gating stabilizes dense-reward RL by rejecting failure-mode rollout groups.}
Rubric-gating ($\mu=1, \rho = 25\%, \nu = 60\%$) consistently complements \rwdname{} on \para{} and \rar{} (Table~\ref{tab:main_results}, rows \#6 vs.\ \#7), improving downstream performance and stabilizing training by rejecting rollout groups whose \emph{entire} answer set fails fundamental task requirements. Without gating, we observe this failure mode in practice: on \para{}, the trained model sometimes stops making any revision while still receiving high dense rewards (Appendix~\ref{app:no_revision}). On \nda{}, the impact is limited because JSON-formatted responses make comprehensive rubric checks hard to express, consistent with the pure rubric baseline (row \#5).

\bbb{Constrained RL's separation of rubric as constraint from reward outperforms merging them into one scalar reward.}
Following the constrained-RL principle of separating optimization from feasibility, \myname{} keeps rubrics out of the reward and uses them only as hard accept/reject gates on rollout groups. The natural alternative is to inject the same rubric judgments as a graded scalar reward added to \rwdname{}, collapsing the two signals into one scalar (Appendix~\ref{app:rubric_reward_vs_gate}). With all other components fixed, this combined-reward variant yields lower performance on both \para{} and \rar{} and additionally falls into the no-revision mode collapse on \para{}, since graded rubric scores still admit gradient updates from high-certainty content-free rollouts that hard gates would reject.

\bbb{Yet such separation drives progressive satisfaction of the feasibility constraints.}
Since rubric judgments are not directly in the reward, a natural concern is whether the policy actually learns to satisfy them or merely slips past the gates on training queries. We measure this two ways (Appendix~\ref{app:heldout_rubric}). (i) \emph{Training-time:} the rubric-gating rejection rate decreases on \para{} and \rar{} (Fig.~\ref{fig:rubric_convergence}), indicating improved \emph{constraint satisfaction} by the policy as training progresses. Note that queries of rejected groups are continuously revisited in later epochs (rejection fractions: $19.6\%/10.4\%/7.2\%$ on \para{}/\rar{}/\nda{}). (ii) \emph{Held-out:} judging unseen test responses against the same query-specific rubrics, \myname{} substantially raises the satisfied fraction on all three datasets, evidencing generalizable constraint satisfaction rather than query-specific memorization.


\vspace{-1.5mm}
\subsection{Effect of Variance-Based Query Filtering}
\label{sec:eval_variance_based_filtering}
\vspace{-1.5mm}

\bbb{Variance-based query filtering speeds and stabilizes convergence.}
The filter consistently improves performance (Table~\ref{tab:main_results}, row \#7 vs.\ \#8 for most datasets; row \#6 vs.\ \#8 for \finqa{}) by skipping ineffective low-variation groups, and yields earlier and more stable convergence (Appendix~\ref{app:stablize_reward}). For our large corpora (tens of thousands of samples, except \nda{}), training on the full set empirically harms final performance~\citep{ye2025limo}; the Section~\ref{sec:variance_filtering} filter retains only \textbf{18\%}, \textbf{16\%}, and \textbf{12.5\%} on \para{}, \rar{}, and \finqa{}, and $200$ of $330$ on \nda{}, underscoring that \emph{quality (high-variance groups) beats quantity}.

\bbb{Token-selective variance outperforms RLPR's all-token average.}
The concurrent RLPR work~\citep{yu2025rlpr} proposes a similar query filter but averages variance over \emph{all} reference tokens. Because uninformative tokens dominate long-form references (the same dilution problem motivating \rwdname{}, Section~\ref{sec:aggregation}), all-token variance is itself diluted, so groups uninformative \emph{on the reasoning-reflective tokens} can still pass the filter. Restricting variance to top-variance tokens (\myname{}'s criterion) better aligns the filter with the actual learning signal and improves over RLPR's filter on top of \rwdname{} on both \para{} and \rar{} (Appendix~\ref{app:rlpr_filter_comparison}).

%% file: 5-conclusion.tex
\vspace{-2mm}
\section{Discussion and Limitations}
\label{sec:limitations}
\vspace{-2mm}
\bbb{Reliance on a reference answer.} \myname{} primarily targets \emph{reference-anchored} tasks without verifiable outcomes but with at least one reasonable-quality reference answer per query. It does not apply to fully reference-free open-ended generation (e.g., creative writing). \rwdname{} also consumes only a single reference per query; extending it to multiple valid references (e.g., averaging \rwdname{} across a reference set) is a natural way to broaden the optimization target and we leave it to future work.

\bbb{Residual reward-hacking risk.} Like all self-certainty-based RL, \rwdname{} can in principle reward a flawed CoT that nonetheless raises reference-token probability. \myname{} mitigates but does not fully eliminate this structural risk shared with RLVR.
\vspace{-2mm}
\section{Conclusion}
\vspace{-2mm}
We introduce \myname{}, a constrained RL recipe for unverifiable tasks with reference answers that cleanly separates optimization from feasibility. \myname{} leverages cross-rollout variance of reference tokens under different reasoning prefixes, both as \rwdname{} that up-weights reasoning-reflective tokens and as a query filter that discards groups with insufficient signal; rubrics, in turn, serve as \emph{gates} on the final answer rather than as rewards. Empirically, across four datasets, \myname{} outperforms strong baselines, learns up to $2$--$3\times$ faster, improves rubric compliance without rubric-derived rewards, and trains more stably and sample-efficiently.


%% file: 6-appendix.tex
\section{GRPO Objective}
\label{app:grpo_objective}
In GRPO~\citep{shao2024deepseekmath}, the surrogate objective (a PPO variant~\citep{schulman2017proximal}) is:
{
\thinmuskip=1mu
\medmuskip=1mu
\thickmuskip=1mu
\begin{align}
\label{eq:grpo}
\mathcal{J}_{\text{GRPO}}(\theta) = & \mathbb{E}_{\textbf{q} \sim P(Q), \{\textbf{o}_i\}_{i=1}^{G} \sim \pi_{\theta_{\text{old}}}(\cdot|\textbf{q})} \nonumber \\
& \Biggl[
\frac{1}{G} \sum_{i=1}^{G} \frac{1}{\lvert\textbf{o}_i\rvert} \sum_{t=1}^{\lvert \textbf{o}_i \rvert}
\left\{ \min \left[
\frac{\pi_{\theta}(o_i | \textbf{q}, \textbf{o}_{i<t})}{\pi_{\theta_{\text{old}}}(o_i | \textbf{q}, \textbf{o}_{i<t})} \hat A_{i,t}, \ 
\text{clip} \left( \frac{\pi_{\theta}(o_i | \textbf{q}, \textbf{o}_{i<t})}{\pi_{\theta_{\text{old}}}(o_i | \textbf{q}, \textbf{o}_{i<t})}, 1 - \epsilon, 1 + \epsilon \right) \hat A_{i, t}
\right] \right\} \nonumber \\
& - \beta \, 
\mathbb{D}_{\text{KL}} \left( \pi_{\theta} \| \pi_{\text{ref}} \right)
\Biggr]
\end{align}
}
where $\epsilon$ is the clipping parameter and $\beta$ weights KL regularization.

\section{GRPO with the self-certainty reward is not equivalent to SFT}
\label{appendix:grpo_self_cert_not_sft}
Let $\mathbf{q}$ be a prompt, $\mathbf{y}^\star$ a fixed reference answer, and $\hat{\mathbf{c}}\sim\pi_\theta(\cdot\mid\mathbf{q})$ a sampled CoT.
Define the self-certainty
$s_\theta(\mathbf{q},\hat{\mathbf{c}};\mathbf{y}^\star)\!=\!\log \pi_\theta(\mathbf{y}^\star\mid \mathbf{q},\hat{\mathbf{c}})$.
GRPO maximizes $\mathcal{J}(\theta)=\mathbb{E}_{\hat{\mathbf{c}}\sim\pi_\theta}[\,s_\theta(\mathbf{q},\hat{\mathbf{c}};\mathbf{y}^\star)\,]$
with score-function gradient (baseline $b$ independent of $\hat{\mathbf{c}}$):
\[
\nabla_\theta \mathcal{J}(\theta)
=\mathbb{E}_{\hat{\mathbf{c}}\sim\pi_\theta}\!\big[(s_\theta(\mathbf{q},\hat{\mathbf{c}};\mathbf{y}^\star)-b(\mathbf{q}))\,\nabla_\theta \log \pi_\theta(\hat{\mathbf{c}}\mid \mathbf{q})\big].
\]

In the standard policy-gradient estimator, $s_\theta(\mathbf{q},\hat{\mathbf{c}};\mathbf{y}^\star)$ is treated as a detached scalar;
hence $\nabla_\theta \mathcal{J}$ has support only on $\nabla_\theta \log \pi_\theta(\hat{\mathbf{c}}\mid \mathbf{q})$, i.e., \emph{CoT-token} scores.
No term of the form $\nabla_\theta \log \pi_\theta(y^\star_j\mid \cdot)$ appears.
By contrast, SFT minimizes
\[
\mathcal{L}_{\text{SFT}}(\theta)
=-\sum_{j=1}^{|\mathbf{y}^\star|} \log \pi_\theta\!\big(y^\star_j \mid \mathbf{q},\mathbf{y}^\star_{<j}\big),
\quad
\nabla_\theta \mathcal{L}_{\text{SFT}}(\theta)
=-\sum_j \nabla_\theta \log \pi_\theta\!\big(y^\star_j \mid \cdot\big),
\]
which places gradients on \emph{answer-token} scores under teacher forcing.
Since the two gradients act on disjoint sufficient statistics (CoT vs.\ answer tokens), they are not equal.
Moreover, even if one augments the trajectory with sampled answers $\mathbf{y}\sim\pi_\theta(\cdot\mid \mathbf{q},\hat{\mathbf{c}})$ while keeping the same reward (independent of $\mathbf{y}$), the answer-token term vanishes by the score-function identity:
\[
\mathbb{E}_{\mathbf{y}}\!\big[\nabla_\theta \log \pi_\theta(\mathbf{y}\mid \mathbf{q},\hat{\mathbf{c}})\big]=0.
\]
Therefore, GRPO with self-certainty does not reduce to SFT. \hfill$\square$

\section{Plain Aggregation of Token-level Self-certainty Diluting the Influence of Informative Tokens in Advantage Score}
\label{appendix:math_analysis_r3}
Let $x_{ij} \in [0,1]$ denote the \emph{token self-certainty} of token $j$ in sample $i$, and assume we have $t$ such self-certainty values per sample. We form the per-sample average
\[
m_i \;=\; \frac{1}{t} \sum_{j=1}^{t} x_{ij}.
\]
Across samples, let $\Sigma$ be the $t \times t$ covariance matrix of the random vector 
\[
\mathbf{x}_i = (x_{i1},\ldots,x_{it})^\top.
\]
Then the variance of the per-sample mean is
\[
\operatorname{Var}(m_i)
\;=\;
\frac{1}{t^2} \, \mathbf{1}^\top \Sigma \, \mathbf{1}.
\]

\paragraph{Independent token self-certainties.}
Assume token self-certainties are independent across $j$, with variances 
$\sigma_1^2, \ldots, \sigma_t^2$. Then
\[
\operatorname{Var}(m_i)
=
\frac{1}{t^2}\sum_{j=1}^{t} \sigma_j^2.
\]
Suppose a single token $j^*$ carries meaningful signal -- e.g., its self-certainty differs systematically across samples -- while the remaining $t-1$ tokens contribute only small random fluctuations with variance $\sigma_0^2$. The contribution of the important token to $m_i$ scales as
\[
\frac{1}{t},
\]
while the aggregate noise variance in the denominator scales as
\[
\frac{1}{t^2}\big(\sigma_{\!*}^2 + (t-1)\sigma_0^2\big)
\;\approx\;
\frac{\sigma_0^2}{t}
\quad\text{as } t \to \infty.
\]
Thus the resulting $z$-score -- essentially the advantage in Eq.~\ref{eq:grpo_advantage} -- behaves as
\[
z_i \;\approx\; \frac{\delta/t}{\sigma_0 / \sqrt{t}}
\;=\;
\frac{\delta}{\sigma_0 \sqrt{t}},
\]
which decays like $1/\sqrt{t}$. Hence, many small self-certainty fluctuations dilute the influence of the single meaningful signal token.

\paragraph{Positively correlated token self-certainties.}
If the $t$ noise tokens have common variance $\sigma_0^2$ and correlation $\rho > 0$, then
\[
\operatorname{Var}(m_i)
=
\frac{1}{t^2}
\left[ t \sigma_0^2 + t(t-1)\rho \sigma_0^2 \right]
=
\sigma_0^2\!\left(\frac{1}{t} + \rho \left(1 - \frac{1}{t}\right)\right).
\]
As $t \to \infty$, this approaches $\rho \sigma_0^2$, i.e., it stops shrinking. But the signal contribution still scales as $1/t$, so the corresponding $z$-score now decays as
\[
z_i \approx \frac{\delta}{\sigma_0 \sqrt{\rho}} \cdot \frac{1}{t},
\]
an even stronger dilution effect. Thus, aggregate noise from many weakly varying token self-certainties dominates the contribution from the important token.

\section{Simulation Setup}
\label{appendix:sim_setup}
We begin by inspecting tens of queries and their corresponding rollout groups across the datasets used in our experiments, manually evaluating the quality of each CoT trace and its final answer, and identifying tokens whose self-certainty exhibits noticeably higher variance across rollouts in alignment with sample quality -- our empirical notion of \emph{reasoning-reflective} tokens. In contrast, the vast majority of tokens display comparatively low variance and provide little discriminative signal. Guided by these observations, we construct a simulation in which each token's self-certainty is modeled as a Gaussian random variable whose mean and standard deviation match representative statistics extracted from our datasets. We further sample the reference outcome length uniformly from $25$ to $750$ tokens, reflecting the range encountered in long-form tasks. For each simulated instance, we generate a group of $16$ rollouts, forming a matrix of size $16 \times \lvert \mathbf{y} \rvert$ containing the token-level self-certainties. For every simulated rollout group, we then compute the correlation between: (i) the advantage scores obtained under the aggregation scheme dictated by the simulation configuration, and (ii) the $z$-scores of the reasoning-reflective tokens. This correlation quantifies the degree to which the aggregate advantage captures the behavior of the informative reasoning-reflective tokens. In our simulation config, we vary (i) the fraction of reasoning-reflective tokens in the reference sequence and (ii) the aggregation rule used to construct the scalar reward from token self-certainties: (\textit{a}) plain average \emph{log}-probability, (\textit{b}) plain average probability, and (\textit{c}) \rwdname{} (introduced in the next section). For each config, we run $5{,}000$ Monte Carlo trials. Note that we simulated the reasoning-reflective tokens to have correlation among themselves for making the correlation scores cleaner. However, in real-world, they typically may not have such correlation. In that case, the plain aggregation will perform even worse.

\section{Rubric-gating Prompts of \myname{}}
\label{appendix:rubric_gen_prompt_dro}
\begin{tcolorbox}[title = Query-specific Rubric Generation System Prompt of \myname{}, breakable, enhanced, sharp corners]
    \textbf{Role:} You are an expert evaluator tasked with generating most essential guardrail rubrics for judging the quality of a model's response to a given query with task description.

    \vspace{0.75em}
    \textbf{Input Format}\\
    You will be given:
    \begin{itemize}
      \item \textbf{Query with Task Description} --- what the user asked the model to do
      \item \textbf{Ground-Truth Response} --- an example of a high-quality answer for that query
    \end{itemize}

    \vspace{0.25em}
    \textbf{Your Objective}\\
    Your job is to infer, from the query and ground-truth response, the \textbf{essential} set of rubrics or criteria that any correct response \textbf{must} satisfy. Your goal is to identify a set of rubrics or criteria that \textbf{must} be satisfied for the response to be considered correct like the ground-truth response. These rubrics will be used as automatic guardrails to catch \textbf{critical errors} in AI generated responses to this query.

    \vspace{0.75em}
    \textbf{Output Format}\\
    Return your rubrics as a JSON array with \textbf{maximum} 10 items (IDs 0--9):
    \begin{tcolorbox}[colback=black!3, sharp corners]
    \ttfamily
    [\\
    \ \ \{ "id": 0, "rubric\_item": "..." \},\\
    \ \ \{ "id": 1, "rubric\_item": "..." \},\\
    \ \ ...\\
    ]
    \end{tcolorbox}

    \vspace{0.25em}
    \textbf{Rubric Item Requirements}
    \begin{enumerate}
      \item \textbf{Binary and checkable}\\
      It must be answerable with Yes or No by inspecting a candidate response.
      \item \textbf{Query and Ground-Truth specific}\\
      It may encode concrete entities, values, steps, or structures that are present in the ground-truth response and are required in a correct response to the given query. \\
      A rubric item \textbf{must} be satisfied by the ground-truth response. \\
      It must \textbf{not} refer to ``the ground-truth'', ``the example answer'', or similar. \\ 
      It must \textbf{not} introduce new constraints not evidenced by the ground-truth response. 
      \item \textbf{Single atomic requirement}\\
      Each rubric item must test exactly one fine-grained property; do not bundle multiple checks into one item.
      \item \textbf{Objective, not subjective}\\
      Do not use vague terms like: good, high-quality, clear, appropriate, useful, detailed unless converted into an explicit, measurable condition. \\
      Ignore purely stylistic or cosmetic traits unless explicitly required by the task.
      \item \textbf{Evaluator-consistent}\\
      Two independent evaluators should reach the same Yes/No decision when applying the rubric.
      \item \textbf{Non-overlapping}\\
      No two rubric items should test the same underlying requirement.
    \end{enumerate}

    \vspace{0.25em}
    \textbf{How to Derive the Rubrics} \\
    You must infer rubric items by analyzing what the Query with Task Description requires and what the Ground-Truth Response demonstrates. Follow these steps:
    \begin{enumerate}
      \item \textbf{Analyze the task:} Identify what makes a response high-quality for this specific query and task.
      \item \textbf{Extract key dimensions from the ground-truth response:} Determine the essential and distinct quality aspects from the ground-truth response that are relevant to the query and task.
      \item \textbf{Prioritize the essential rubrics:} Focus on the important and essential aspects of response quality. Keep only rubrics that would catch critical errors like guardrails in the AI generated responses to this query.
      \item \textbf{Validate requirements compliance:} Make sure each rubric satisfies all the requirements stated in \textit{Rubric Item Requirements}.
      \item \textbf{Non-Conflicting (Global Consistency Check):} Ensure that the ground-truth response satisfies all rubric items simultaneously. Do not include items that contradict each other.
    \end{enumerate}
\end{tcolorbox}

\begin{tcolorbox}[title = Rubric Judgment System Prompt of \myname{}, breakable, enhanced, sharp corners]

\textbf{Role:} You are an expert evaluator tasked with objectively assessing an AI-generated response against a set of quality rubrics.

\vspace{0.75em}
\textbf{Input Format}\\
You will be given:
\begin{itemize}
    \item \textbf{Query with Task Description} — what the user asked the model to do
    \item \textbf{Model Response} — the AI-generated response to evaluate
    \item \textbf{Rubric JSON} — a list of binary quality criteria
\end{itemize}

\vspace{0.75em}
\textbf{Your Objective}\\
Evaluate the Model Response against each rubric item independently and determine:
\begin{itemize}
    \item Whether each rubric is satisfied (Yes/No)
    \item A brief justification for each judgment
\end{itemize}

\vspace{0.75em}
\textbf{Evaluation Guidelines}

\textbf{Core Principles}
\begin{itemize}
    \item \textbf{Independence:} Evaluate each rubric item separately without letting other criteria influence your judgment.
    \item \textbf{Objectivity:} Base decisions solely on observable evidence in the Model Response.
    \item \textbf{Binary Decisions:} Every rubric must receive exactly ``Yes'' or ``No'' (no partial credit).
    \item \textbf{Strict Interpretation:} When in doubt, default to the most objective reading of the criterion.
    \item \textbf{Response-Only Focus:} Evaluate only what's in the Model Response.
\end{itemize}

\textbf{Decision Framework}
\begin{itemize}
    \item \textbf{Answer ``Yes'' when:}
    \begin{itemize}
        \item The criterion is clearly and unambiguously satisfied.
        \item Observable evidence directly supports the rubric requirement.
        \item Any reasonable evaluator would reach the same conclusion.
    \end{itemize}

    \item \textbf{Answer ``No'' when:}
    \begin{itemize}
        \item The criterion is not met.
        \item Evidence is ambiguous or insufficient.
        \item The response only partially satisfies the requirement.
        \item You are uncertain (err on the side of ``No'' for binary clarity).
    \end{itemize}
\end{itemize}

\textbf{Common Pitfalls to Avoid}
\begin{itemize}
    \item Do not be lenient due to overall response quality.
    \item Do not infer intent; evaluate only what is written.
    \item Do not let earlier rubric judgments bias later ones.
    \item Do not give credit for “almost” satisfying a criterion.
\end{itemize}

\vspace{0.75em}
\textbf{Evaluation Process}

For each rubric item:
\begin{itemize}
    \item Read the rubric carefully: understand exactly what it's asking.
    \item Examine the Model Response: look for specific evidence.
    \item Make a binary decision: Yes or No.
    \item Document your reasoning: cite specific evidence or explain the gap.
    \item Move to the next rubric: reset your judgment framework.
\end{itemize}

\vspace{0.75em}
\textbf{Output Format}\\
Return your evaluation as a JSON array in the following structure:

\begin{tcolorbox}[colback=black!3, sharp corners]
\ttfamily
[\\
\ \ \{ "id": 0, "satisfied": true, "justification": "Brief justification of the judgement" \},\\
\ \ \{ "id": 1, "satisfied": false, "justification": "Brief justification of the judgement" \},\\
\ \ ...\\
]
\end{tcolorbox}
\end{tcolorbox}

\section{Evaluation Setup Details}
\label{app:eval_setup}\subsection{Datasets and Metrics}
\label{app:data_stats}
\label{sec:datasets}
We use four datasets spanning diverse domains and tasks. 

\textbf{\para{}}~\citep{jourdan2025pararev} contains 48K original-revised paragraph pairs with reviews; we focus on the initial revision and extend each sample with preceding/following paper context by locating paragraphs in the raw papers and extracting context from CASIMIR~\citep{jourdan2024casimir}, yielding an adapted 12K-sample set (80\%/20\% train/test). The task is paragraph revision conditioned on paper context and reviewer feedback; we evaluate with pairwise win rate using LLM judges and the prompt template in Appendix~\ref{appendix:win_rate}.  

\textbf{\rar{}}~\citep{gunjal2025rubrics} provides 20K rubric-annotated medicine-domain prompts with reference answers; the task is medical QA, and we compute rubric-supervised pairwise win rate by having judges select the response that better satisfies the provided rubric, using the prompt template in Appendix~\ref{appendix:win_rate}.


\textbf{\nda{}}~\citep{koreeda2021contractnli} contains 607 annotated contracts paired with 17 fixed hypotheses, where the task is to predict whether each hypothesis is entailed, contradicted, or not mentioned. We treat the output as a JSON schema containing the 17 predictions and the corresponding supporting evidence, and report macro-F1 across the three labels.

\textbf{\finqa{}}~\citep{chen2021finqa} comprises 8K+ finance QA instances; the task is numerical reasoning over financial context, and we score answer correctness with math-verify~\citep{mathverify2025} under a 0.001 tolerance.
Detailed dataset statistics are summarized in Table~\ref{tab:dataset_stats}.
Except for \finqa{} (primarily numerical reasoning), all benchmarks involve long-context inputs (query plus substantial supporting context) and non-trivial outputs (paragraph-level text or structured JSON).

\begin{table*}[h]
\centering
\caption{Dataset statistics. Input length denotes tokens in query plus context; output length denotes tokens in the target outcome. Tokenization uses the Qwen3 tokenizer.}
\label{tab:dataset_stats}
\small
\setlength{\tabcolsep}{2pt}
\begin{tabular}{lcccc}
\toprule
 & \textbf{\para{}} & \textbf{\rar{}}  & \textbf{\nda{}} & \textbf{\finqa{}} \\
\midrule
\textbf{Domain} & Scientific writing & Medicine  & Legal contracts & Finance \\
\textbf{\# Samples} & 12K & 20K  & 330 & 8,281 \\
\textbf{Avg. Input Len (tokens)} & 4212 & 205  & 3815 & 1158 \\
\textbf{Avg. Output Len (tokens)} & 258 & 116  & 1204 & 6 \\
\textbf{Metric} & Pairwise win rate & Rubric win rate  & Macro-F1 & Math-verify (0.001 tol.) \\
\bottomrule
\end{tabular}
\end{table*}

For LLM-based evaluations, we use OpenAI o4-mini~\cite{o4mini} as the evaluator. To control for positional bias, we repeat pairwise judgments in both response orders and report the averaged scores.
Unless otherwise specified, we follow the original splits for each dataset, training a separate model on the training split and evaluating it on the corresponding test split.

\subsection{Training and Inference Settings.}
We conduct \myname{} training using Qwen3-14B~\citep{yang2025qwen3} on all datasets except \finqa{}, where we use DeepSeek-R1-Distill-Qwen-7B~\citep{guo2025deepseek}. 
Each training step processes a batch of 16 samples, and the actor model generates 16 responses per question with temperature = 1.0 and top\_p = 0.95. 
We use a learning rate of $5.0\times e^{-7}$, a warmup ratio of 0.2, and a constant-with-warmup scheduler, selecting checkpoints based on the best validation reward. 
For GRPO optimization, we adopt the loss from \citet{liu2025understanding}, with scaled rewards, masking for truncated completions, and an upper clipping coefficient $\epsilon_{\text{high}} = 0.2$. 
Following \citet{wang2025beyond}, we additionally restrict policy-gradient updates to high-entropy CoT tokens (the ``80/20'' rule), keeping the top-$\rho$ fraction of CoT tokens by per-step entropy; we use $\rho \in [0.3, 0.5]$ across datasets, which we found to give a good balance between training stability and exploration.
Although prior work typically sets entropy regularization $\beta=0$, we find $\beta=0.001$ improves stability and convergence. Following prior empirical findings~\citep{liu2025understanding}, we do not use a chat template during training.
Training runs on three nodes with $8 \times$ NVIDIA A100 GPUs each, using HuggingFace TRL for reinforcement learning, DeepSpeed for distributed training, and vLLM for rollout generation, \rwdname{} computation, and rubric generation and judgment. 
During inference, we use temperature = 0.6 and top\_p = 0.95, and report average numbers across all test samples.

\subsection{\myname{}-Specific Hyperparameters}
\label{app:hyperparam_table}
Table~\ref{tab:hparams} consolidates the \myname{}-specific hyperparameters.

\begin{table}[h]
\centering
\caption{Consolidated \myname{}-specific hyperparameters.}
\label{tab:hparams}
\small
\setlength{\tabcolsep}{2pt}
\renewcommand{\arraystretch}{1.2}
\begin{tabular}{l l l}
\toprule
\textbf{Component} & \textbf{Hyperparameter} & \textbf{Value} \\
\midrule
\multirow{3}{*}{\rwdname{}}
 & Emphasis $\omega$       & $1$--$2$ (per dataset) \\
 & Lower clip $\lambda_{\text{low}}$  & $\lesssim 0.05$ \\
 & Upper clip $\lambda_{\text{high}}$ & $\gtrsim 0.85$ \\
\midrule
\multirow{3}{*}{Rubric-gating}
 & Coverage $\mu$    & $1$ \\
 & Top-rank band $\rho$   & $25\%$ \\
 & Consistency $\nu$    & $60\%$ \\
\midrule
\multirow{2}{*}{Variance filter}
 & Top fraction $q$ & $10\%$ \\
 & Retention (\para{} / \rar{} / \nda{} / \finqa{}) & $18\%$ / $16\%$ / $60.6\%$ / $12.5\%$ \\
\bottomrule
\end{tabular}
\end{table}

\subsection{Training and Inference Efficiency}
\label{app:efficiency}

We report training- and inference-time costs for \myname{} alongside the strongest self-certainty baseline (Avg Prob, Table~\ref{tab:main_results} row \#4) and, where applicable, the base model. All numbers are measured under the same hardware setup as Appendix~\ref{app:eval_setup} (3 nodes $\times$ 8 NVIDIA A100 GPUs). \myname{} matches Avg Prob in per-step training throughput and peak GPU memory, while reaching comparable downstream performance in substantially fewer total GPU-hours due to variance-based filtering (Sections~\ref{sec:eval_variance_based_filtering} and~\ref{sec:eval:r3}). Average completion length at inference is also comparable across methods, indicating no inference-time overhead from \myname{}.

\begin{table}[h]
\centering
\caption{Training and inference efficiency for \myname{} vs.\ the Avg Prob baseline (Table~\ref{tab:main_results} row \#4) and the base model. Per-step throughput and peak memory are comparable, while \myname{} reaches similar downstream performance in $1.5$--$2.7\times$ fewer total GPU-hours, with no inference-time overhead.}
\label{tab:efficiency}
\small
\setlength{\tabcolsep}{4pt}
\renewcommand{\arraystretch}{1.15}
\begin{tabular}{lccc}
\toprule
\textbf{Metric (\para{} / \rar{} / \nda{})} & \textbf{\myname{}} & \textbf{Avg Prob} & \textbf{Base} \\
\midrule
Training throughput (samples/hr)                       & 108 / 135 / 112 & 122 / 141 / 130 & -- \\
Peak GPU memory (GB per node)                          & 78.6 / 79.1 / 78.4 & 78.4 / 79.4 / 78.2 & -- \\
Total GPU-hours to reach similar performance           & \textbf{370 / 382 / 624} & 1008 / 728 / 945 & -- \\
Avg completion tokens at inference (CoT + answer)      & 1.9k / 1.2k / 1.9k & 1.7k / 1.2k / 1.8k & 0.9k / 0.8k / 2.7k \\
\bottomrule
\end{tabular}
\end{table}

\section{Evaluation Baseline: RL from Rubrics as Rewards}
\label{appendix:rubric_baseline}
We include a rubric-aggregated reward baseline that optimizes the policy, with GRPO objective, using a scalar score obtained by evaluating each rollout (final answer only) against query-specific rubrics and averaging across all rubrics. This setup is inspired by two contemporaneous lines of work -- \emph{Rubrics as Rewards} (RaR)~\cite{gunjal2025rubrics} and \emph{RL with Evolving Rubrics} (RLER)~\cite{shao2025dr} -- both of which employ query-specific positive and negative rubrics to score responses, with RLER additionally generating \emph{evolving} (discriminative) rubrics on the fly to keep pace with model improvements during training. In our implementation, we first construct \emph{static}, gradable rubrics for each query using the query and its reference answer using GPT-5 model (prompt is provided below). These rubrics are not yes/no checks but gradable positive and negative criteria. During RL, we further instantiate \emph{evolving} rubrics per query using the RLER procedure and prompt to elicit discriminative criteria from the model's current response set. Each response is then scored by an LLM-as-judge (GPT-4o) on a \([0,2]\) scale for every rubric (positive and negative), using the judge prompt from RLER, and we aggregate by averaging scores across all rubrics associated with that query to produce a single scalar reward for policy optimization.

\begin{tcolorbox}[title= Query-specific Gradable Rubrics Generation Prompt, breakable, enhanced, sharp corners]

You are an expert evaluator tasked with generating the most essential guardrail rubrics for judging the quality of a model's response to a given query with task description.

\section*{Input Format}
You will be given:
\begin{itemize}
    \item \textbf{Query with Task Description} — what the user asked the model to do
    \item \textbf{Ground-Truth Response} — an example of a high-quality answer for that query
\end{itemize}

\section*{Objective}

Your job is to infer, from the query and ground-truth response, the \textbf{essential} set of rubrics or criteria that any correct response must satisfy. These rubrics represent the most important quality dimensions and failure modes determining correctness for this task.

You must produce two categories:
\begin{enumerate}
    \item \textbf{Positive Rubrics}: critical quality aspects demonstrated by the ground-truth response
    \item \textbf{Negative Rubrics}: critical flaws that definitively degrade response quality for this task
\end{enumerate}

The rubrics should be phrased as \textbf{gradable distinctions}, describing what strong performance looks like (positive rubrics) and what severe failure looks like (negative rubrics).  
They must not include scores or scoring instructions.

\section*{Output Format}

Return your rubrics in the following JSON structure:

\begin{tcolorbox}[colback=black!3, sharp corners]
\ttfamily
\{\\
\ \ "positive\_rubrics": [\\
\ \ \ \ \{ "description": "<detailed excellence description>", "title": "<abstract label>" \}\\
\ \ ],\\[2pt]
\ \ "negative\_rubrics": [\\
\ \ \ \ \{ "description": "<detailed failure description>", "title": "<abstract label>" \}\\
\ \ ]\\
\}
\end{tcolorbox}

\section*{Rubric Item Requirements}

Each rubric item must satisfy all of the following:

\begin{enumerate}
    \item \textbf{Gradable and inspectable}  
    Describes a clear evaluative distinction that can be judged in a candidate response; not phrased as a Yes/No question.

    \item \textbf{Query and Ground-Truth specific}  
    Encodes concrete entities, values, steps, or structures supported by the ground-truth response.  
    Positive rubrics must be exemplified by the ground-truth response.  
    Negative rubrics must correspond to concrete harmful failure modes.  
    Must not reference ``ground-truth'' or ``example answer.''  
    Must not introduce constraints unsupported by the ground-truth response.

    \item \textbf{Single atomic requirement}  
    Represents exactly one fine-grained property or failure mode.

    \item \textbf{Objective}  
    Avoid vague terms such as ``good'' or ``appropriate'' unless they are made explicit and concrete.

    \item \textbf{Evaluator-consistent}  
    Independent evaluators should reach the same judgment on whether the response satisfies the positive rubric or exhibits the negative rubric.

    \item \textbf{Non-overlapping}  
    No two rubrics should describe the same underlying requirement.
\end{enumerate}

\section*{How to Derive the Rubrics}

You must infer rubric items by analyzing what the Query with Task Description requires and what the Ground-Truth Response demonstrates. Follow these steps:

\begin{enumerate}
    \item Analyze the task: identify what makes a response high-quality.
    \item Extract key dimensions from the ground-truth response.
    \item Extract key failure modes: identify critical errors or omissions that make a response unacceptable.
    \item Prioritize essentials: keep only rubrics representing critical quality aspects or critical flaws.
    \item Perform a global consistency check: ensure the ground-truth satisfies all positive rubrics and none of the negative rubrics contradict them.
\end{enumerate}

\end{tcolorbox}

\section{Prompt Templates}
\label{appendix:win_rate}

\label{app:prompt}

\begin{tcolorbox}[
  title={Pairwise Win Rate Prompt for \para{}},
  breakable,
  enhanced,
  sharp corners
]

\section*{Task}

You are an expert in scientific writing and are evaluating two candidate revisions of a paragraph in an academic paper. You will be provided with:
\begin{itemize}
  \item \textbf{Paper Context:} surrounding content of the paragraph
  \item \textbf{Reviewer Comments:} feedback from peer reviewers that may motivate the revision
  \item \textbf{Paragraph to Revise:} the original paragraph to revise
  \item \textbf{Golden Revision:} an expert-written revision serving as the reference
  \item \textbf{Candidate Revisions:} two candidate revisions to evaluate
\end{itemize}
Your goal is to identify the changes made in the golden revision and determine which candidate revision best reflects those changes.

\medskip
\textbf{Paper Context}

\{paper\_context\}

\medskip
\textbf{Reviewer Comments}

\{reviewer\_comments\}

\medskip
\textbf{Paragraph to Revise}

\{paragraph\_to\_revise\}

\medskip
\textbf{Golden Revision}

\{golden\_revision\}

\medskip
\textbf{Candidate Revisions}

\begin{itemize}
  \item Revision A: \{revision\_a\}
  \item Revision B: \{revision\_b\}
\end{itemize}

\medskip
\section*{Instructions}

\begin{enumerate}
  \item \textbf{Understand the context:} Carefully read the \emph{Paper Context} and \emph{Reviewer Comments} to understand the background of the paper and the concerns raised by the reviewers.
  \item \textbf{Identify the desired changes:} Compare the \emph{Golden Revision} with the \emph{Paragraph to Revise}.
  \begin{itemize}
    \item Identify and itemize all changes in content, structure, and language.
    \item Reason how each change addresses specific reviewer comments.
  \end{itemize}
  \item \textbf{Evaluate the Candidate Revisions:} Analyze \emph{Revision A} and \emph{Revision B} by comparing them to the \emph{Paragraph to Revise}, and determine which candidate revision best reflects the changes made in the golden revision. Focus on the following criteria:
  \begin{itemize}
    \item \textbf{Effectiveness:} How well does each candidate revision address the specific reviewer comments, following the approach demonstrated by the golden revision?
    \item \textbf{Precision:} How well does each candidate revision avoid unnecessary changes and verbose elaboration not present in the golden revision?
  \end{itemize}
  \item \textbf{Make your judgment:} Provide brief reasoning following the steps above, followed by your final decision in the format below:
\end{enumerate}

\begin{tcolorbox}[colback=black!3, sharp corners]
\ttfamily
Reasoning: <Your brief reasoning>

Judgment: <"Revision A" or "Revision B">
\end{tcolorbox}

\end{tcolorbox}

\begin{tcolorbox}[
  title={Rubric-Supervised Win Rate Prompt for \rar{}},
  breakable,
  enhanced,
  sharp corners
]

\section*{Task}

You are an expert evaluator for medical question answering. You will be provided with:
\begin{itemize}
  \item \textbf{Query:} a medical QA prompt
  \item \textbf{Reference Answer:} an expert-written answer
  \item \textbf{Rubric:} a list of criteria, each with a title, description, and weight
  \item \textbf{Candidate Answers:} two candidate answers to evaluate
\end{itemize}

Your goal is to determine which candidate better answers the query based on the rubric.

\medskip
\textbf{Query}

\{query\}

\medskip
\textbf{Reference Answer}

\{reference\_answer\}

\medskip
\textbf{Rubric}

\{rubric\}

\medskip
\textbf{Candidate Answers}

\begin{itemize}
  \item Answer A: \{answer\_a\}
  \item Answer B: \{answer\_b\}
\end{itemize}

\medskip
\section*{Instructions}

\begin{enumerate}
  \item \textbf{Understand the query:} Carefully read the \emph{Query} to understand the medical context, what is being asked, and any constraints. Read the \emph{Reference Answer} to understand what a high-quality answer looks like.

  \item \textbf{Evaluate each candidate using the rubric:} For \emph{each} rubric item, compare \emph{Answer A} with \emph{Answer B} and determine which better satisfies the criterion. Aggregate the comparisons across all rubric items by their weights to determine which candidate better satisfies the rubric overall.

  \item \textbf{Output Format:} Reason step-by-step, and respond with \textbf{only} a JSON object containing your step-by-step reasoning and final judgment in the format below:
\end{enumerate}

\begin{tcblisting}{
  colback=black!3,
  sharp corners,
  listing only,
  breakable,
  listing options={
    basicstyle=\ttfamily,
    breaklines=true,
    columns=fullflexible
  }
}
{
  "reasoning": "Your step-by-step reasoning.",
  "judgment": "'Answer A' or 'Answer B'. If both are similarly good, choose one based on your best judgment."
}
\end{tcblisting}

\end{tcolorbox}

\section{Lexical Similarity of \myname{} Outputs to the Reference Answer}
\label{app:rouge_l}

As an additional check, we report ROUGE-L F1 between final outputs and the reference answer for the base model, the lexical-similarity baseline (RL-F1, Table~\ref{tab:main_results} row \#2), and full \myname{} (row \#8) in Table~\ref{tab:rouge_l}. Across the three long-form datasets, \myname{} attains substantially higher downstream win rates than the base model with only modest increases in ROUGE-L, and remains far below the explicit lexical-similarity baseline. This is consistent with the design of \rwdname{}, which rewards CoTs that raise the model's likelihood of the reference answer through autoregressive conditioning (Appendix~\ref{appendix:grpo_self_cert_not_sft}) rather than directly targeting surface-form similarity.

\begin{table}[h]
\centering
\caption{ROUGE-L F1 (\%) of final outputs against the reference answer. \myname{} achieves the highest downstream win rates (Table~\ref{tab:main_results}) without becoming substantially more lexically similar to the reference, in contrast to the lexical-similarity baseline RL-F1. \finqa{} is omitted because its short numeric answers make ROUGE-L uninformative.}
\label{tab:rouge_l}
\small
\setlength{\tabcolsep}{8pt}
\renewcommand{\arraystretch}{1.15}
\begin{tabular}{lccc}
\toprule
\textbf{Method (Row \# in Table~\ref{tab:main_results})} & \textbf{\para{}} & \textbf{\rar{}} & \textbf{\nda{}} \\
\midrule
Base (\#1)              & 56.4 & 18.5 & 38.1 \\
RL-F1 (\#2)             & 77.2 & 26.9 & 76.4 \\
\myname{} (\#8)         & 61.7 & 22.1 & 46.2 \\
\bottomrule
\end{tabular}
\end{table}

\section{Comparison with SFT and On-Policy Distillation}
\label{app:sft_distillation}

A natural question is whether \myname{}'s gains really stem from RL formulation, since \myname{} leverages a reference answer for both \rwdname{} computation and rubric construction. To address this, we compare against two natural non-RL baselines that also rely on reference supervision: (i) supervised fine-tuning (SFT) on the reference answers and (ii) on-policy distillation from a stronger teacher model.

For SFT, we fine-tune the same Qwen3-14B base model on the training-set reference answers using the standard cross-entropy objective. For on-policy distillation, we follow GKD~\citep{agarwal2024policy} with Qwen3-235B-A22B as the teacher and Qwen3-14B as the student, distilling on the same training prompts. All methods are evaluated under the same protocol as Table~\ref{tab:main_results} (win rate vs.\ the base model).

Table~\ref{tab:sft_distillation} reports the results. SFT clearly \emph{degrades} performance on \para{} and is roughly neutral on \rar{}, indicating that direct imitation of a single reference answer fails to extract useful improvement signal and can even harm the base model in these settings. On-policy distillation from a substantially stronger teacher improves over the base model but still trails \myname{} by a wide margin, despite using a much larger teacher. These results indicate that \myname{}'s gains do not come merely from access to a reference answer: the RL formulation extracts substantially more learning signal from the same reference than SFT or distillation can.

\begin{table}[h]
\centering
\caption{Comparison with SFT and on-policy distillation baselines that also use reference supervision. Student model is Qwen3-14B in all rows. Win rate (\%) vs.\ the base model.}
\label{tab:sft_distillation}
\small
\setlength{\tabcolsep}{8pt}
\renewcommand{\arraystretch}{1.15}
\begin{tabular}{lcc}
\toprule
\textbf{Method} & \textbf{\para{}} & \textbf{\rar{}} \\
\midrule
Base                                                          & 50.0 & 50.0 \\
SFT                                    & 43.1 & 49.7 \\
On-policy distillation        & 53.6 & 53.1 \\
Full \myname{}                                                & \textbf{63.7} & \textbf{57.6} \\
\bottomrule
\end{tabular}
\end{table}

\section{Transferability to External Benchmarks}
\label{app:healthbench}
We evaluate the transferability of \myname{} by testing checkpoints trained on \rar{} on the external HealthBench benchmark \cite{arora2025healthbench} for both Qwen3-4B and Qwen3-14B. HealthBench is also a medical question-answering benchmark, but it is constructed independently from \rar{}, making it suitable for assessing cross-dataset generalization. \myname{} improves Qwen3-4B from 42.8 to 44.9 (+2.1) and Qwen3-14B from 48.7 to 49.3 (+0.6). The smaller gain for Qwen3-14B is expected, as the base model is already strong (comparable to GPT-4.1), leaving less headroom for improvement from \rar{}. These results suggest that \myname{} encourages generalizable reasoning rather than merely memorizing dataset-specific patterns.

\section{Robustness to Reference Quality}
\label{app:imperfect_ref}

To evaluate \myname{}'s sensitivity to reference quality, we replace the human-written ground-truth (GT) reference with a model-generated one (best-of-4 from Claude Sonnet~4.6, judged by the same evaluator as in win-rate scoring) to simulate an imperfect, off-distribution reference, and re-train \myname{} with all other components unchanged. As shown in Table~\ref{tab:imperfect_ref}, the model-reference variant still improves over the base on both \para{} and \rar{}, with smaller magnitudes than under the GT reference. Reference quality therefore affects the magnitude, not the existence, of the improvement, indicating that \myname{} remains usable in settings where only imperfect references are available.

\begin{table}[h]
  \centering
  \caption{\textbf{Robustness to reference quality.} Win rate (\%) vs.\ the base model on \para{} and \rar{} (Qwen3-14B). \myname{} continues to improve over the base even when trained with an imperfect, model-generated reference.}
  \label{tab:imperfect_ref}
  \small
  \setlength{\tabcolsep}{8pt}
  \begin{tabular}{lcc}
    \toprule
    \textbf{Setting} & \textbf{\para{}} & \textbf{\rar{}} \\
    \midrule
    Claude reference vs.\ Base                  & 58.9 & 56.5 \\
    \myname{} with Claude reference vs.\ Base   & 57.4 & 55.1 \\
    \myname{} with GT reference vs.\ Base       & 63.7 & 57.6 \\
    \bottomrule
  \end{tabular}
\end{table}

\section{Ablation: \rwdname{} Without CoT/Answer Slicing}
\label{app:no_slicing}

In the main \myname{} pipeline, we replace the sampled final answer with the reference answer when computing \rwdname{}, so that the prefix conditioning the reference-answer likelihood is the sampled CoT alone. This slicing relies on a model-provided boundary between CoT and final answer (e.g., \verb|<think>| tags or a structured answer delimiter such as \verb|\boxed{}|). To assess whether \rwdname{} depends on this separation, we ablate slicing by using the \emph{full sampled completion} (CoT + sampled answer) as the prefix for the reference-answer likelihood, leaving everything else unchanged. For a clean comparison, we report \rwdname{} alone here, without rubric-gating or variance-based filtering.

Table~\ref{tab:no_slicing} shows that \rwdname{} with and without slicing perform comparably on both \para{} and \rar{}. This empirically supports the view that \rwdname{} measures how much the sampled prefix raises the likelihood of the correct answer overall, rather than relying on a strict lexical boundary between CoT and final answer. Slicing remains useful operationally -- it preserves the model's learned ``reasoning then answer'' generation pattern so that the final answer can be reliably extracted at inference time -- but it is not a precondition for \rwdname{} to be informative. In particular, this also indicates that \rwdname{} can be applied to models without explicit CoT delimiters.

\begin{table}[h]
\centering
\caption{Ablation on CoT/answer slicing in \rwdname{}. ``Without slicing'' uses the full sampled completion (CoT + sampled answer) as the prefix when scoring the reference answer; ``with slicing'' uses only the sampled CoT. Both variants disable rubric-gating and variance-based filtering for a clean comparison. Win rate (\%) vs.\ the base model.}
\label{tab:no_slicing}
\small
\setlength{\tabcolsep}{8pt}
\renewcommand{\arraystretch}{1.15}
\begin{tabular}{lcc}
\toprule
\textbf{Method} & \textbf{\para{}} & \textbf{\rar{}} \\
\midrule
Base (Table~\ref{tab:main_results}, row \#1)                 & 50.0 & 50.0 \\
\rwdname{} with slicing (Table~\ref{tab:main_results}, row \#6) & 57.8 & 54.2 \\
\rwdname{} without slicing                                  & 57.5 & 54.3 \\
\bottomrule
\end{tabular}
\end{table}

\section{Faster Learning with \rwdname{}}
\label{app:r3_faster_learning}

Table~\ref{tab:r3_faster_learning} reports the number of training steps required for (i) plain Avg Prob to reach its peak downstream performance and (ii) \rwdname{} to reach the same level. \rwdname{} reaches comparable downstream performance in $2$-$3{\times}$ fewer training steps on \para{}, \rar{}, and \nda{}; \finqa{} shows a smaller gap due to short answers and limited dilution. Per-step throughput and peak memory are comparable across methods (Appendix~\ref{app:efficiency}), so step savings translate to wall-clock savings. For fairness, we disable rubric-gating and variance-based filtering for both methods. This induces late-stage performance collapse, slightly earlier for \rwdname{}, as its comparative weighting can amplify groupwise $z$-scores when within-group reward variation is low—motivating the variance-based filter (Section~\ref{sec:variance_filtering}).

\begin{table}[h]
  \centering
  \caption{\textbf{Accelerated learning with \rwdname{}.} Training steps required for Avg Prob to reach its peak downstream performance and for \rwdname{} to reach the same performance level (\emph{lower is better}).}
  \label{tab:r3_faster_learning}
  \setlength{\tabcolsep}{8pt}
  \begin{tabular}{lcc}
    \toprule
    \multirow{2}{*}{\textbf{Dataset}} &
    \multicolumn{2}{c}{\begin{tabular}[c]{@{}c@{}}
      \textbf{\# Steps to Reach Similar} \\
      \textbf{Downstream Performance}
    \end{tabular}} \\
    \cmidrule(lr){2-3}
    & \textbf{Avg Prob} & \textbf{\rwdname{}} \\
    \midrule
    \para{}      & 288 & 96  \\
    \rar{}       & 256 & 128 \\
    \nda{}       & 320 & 208 \\
    \finqa{}     & 384 & 320 \\
    \bottomrule
  \end{tabular}
\end{table}

\section{Ablation on Reasoning Reflectivity Emphasis Factor $\omega$ and Clipping}
\label{app:omega_clipping}

We ablate two key components of \rwdname: (1) the reasoning reflectivity emphasis factor $\omega$, which upweights reasoning-reflective tokens in the reward, and (2) reward clipping described in Section~\ref{sec:r3_design}. We compare three settings: \textit{(i)} plain Avg Prob reward without reasoning reflectivity emphasis ($\omega{=}0$) or clipping, \textit{(ii)} \rwdname{} with clipping but no emphasis ($\omega{=}0$), and \textit{(iii)} the full \rwdname{} with $\omega{>}0$ (dataset-specific, as clarified in Section~\ref{sec:eval:r3}) and clipping enabled.

Table~\ref{tab:omega_clip_ablation} shows that clipping alone provides consistent gains over Avg Prob, indicating its stabilizing role. Adding reflectivity emphasis ($\omega{>}0$) further improves performance across domains, with the largest gains on long-form generation tasks (\para{}, \rar{}), suggesting that explicitly rewarding reflective reasoning behavior is particularly beneficial for long-form and structured generation.

\begin{table}[h]
\centering
\caption{Ablation of reasoning reflectivity emphasis factor $\omega$ and reward clipping. Metrics follow the main evaluation protocol.}
\label{tab:omega_clip_ablation}
\small
\begin{tabular}{lcccc}
\toprule
Method & \para{} & \rar{} & \nda{} & \finqa{} \\
\midrule
Avg Prob ($\omega{=}0$, no clip) & 54.3 & 52.3 & 78.2 & 66.8 \\
\rwdname ($\omega{=}0$, clip) & 55.1 & 53.4 & 79.2 & 66.7 \\
\rwdname ($\omega{>}0$, clip) & \textbf{57.8} & \textbf{54.2} & \textbf{80.6} & \textbf{67.1} \\
\bottomrule
\end{tabular}
\end{table}

\section{No-Revision Mode Collapse on \para{} Without Rubric Supervision}
\label{app:no_revision}

\begin{figure}[t]
    \centering
    \includegraphics[width=0.4\linewidth]{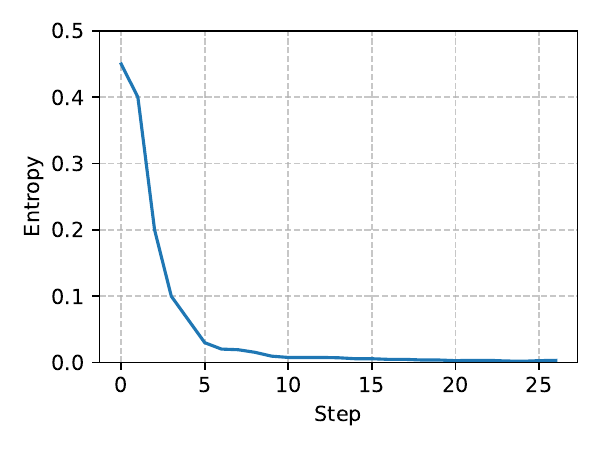}
    \caption{Policy entropy collapse during training on \para{} without rubric supervision.}
    \label{fig:no_revision_entropy}
\end{figure}

On \para{}, training with dense reward alone can lead to a degenerate equilibrium in which the model collapses to a \emph{no-revision} policy while still receiving high reward. Figure~\ref{fig:no_revision_entropy} shows a representative \para{} run where the model converges to making no revisions. The policy entropy quickly falls to near zero after a brief exploratory phase, reflecting collapse of exploration. When effective edits are not found, avoiding changes maximizes self-certainty under the dense reward, producing a stable but unproductive local optimum.

Rubric supervision mitigates this collapse by excluding rollout groups where all responses fail basic task criteria. This prevents high-certainty but content-free no-revision trajectories from contributing to updates, ensuring learning remains driven by minimally valid revisions.

\section{Rubric-as-Reward vs.\ Rubric-as-Gate within \myname{}}
\label{app:rubric_reward_vs_gate}

A natural alternative to gating is to inject the same rubric judgments as a graded scalar reward added to \rwdname{}, rather than using them as hard accept/reject constraints at the rollout-group level. To isolate this design choice, we compare two variants that share the same \rwdname{} dense reward and differ only in how rubric supervision is incorporated:
\begin{itemize}[leftmargin=*,topsep=2pt,itemsep=2pt,parsep=2pt]
  \item \textbf{\rwdname{} + Rubric reward}: rubric judgments are converted into a graded scalar score (using the same query-specific rubric construction and judge as in Appendix~\ref{appendix:rubric_baseline}); both the rubric score and the \rwdname{} reward are normalized to $[0,1]$ and combined as an equally weighted sum (i.e., $0.5\cdot\text{Rubric} + 0.5\cdot\text{\rwdname{}}$); no gating is applied.
  \item \textbf{\rwdname{} + Rubric-gating (\myname{})}: identical rubric judgments are instead used only as hard accept/reject checks on rollout groups, exactly as defined in Section~\ref{sec:eval_rubric_gating}.
\end{itemize}

Table~\ref{tab:rubric_reward_vs_gate} reports the comparison on \para{} and \rar{}. For reference, we also re-list two single-signal baselines from Table~\ref{tab:main_results}: the pure rubric-graded reward (row \#5), which uses the same rubric construction and judge but no \rwdname{} signal, and \rwdname{} alone (row \#6), which uses no rubric supervision. Within an otherwise identical \myname{} pipeline, rubric-as-reward (with or without \rwdname{}) remains substantially weaker than rubric-as-gate, and combining rubric-as-reward with \rwdname{} does not improve over either single-signal baseline. On \para{}, rubric-as-reward additionally fails to prevent the same \emph{no-revision} mode collapse documented in Section~\ref{sec:eval_rubric_gating} and Appendix~\ref{app:no_revision}: graded rubric scores still admit gradient updates from high-certainty content-free rollouts, whereas rubric-gating excludes such groups outright. This directly validates the design choice of treating rubrics as feasibility constraints rather than as dense reward signals.

\begin{table}[h]
\centering
\caption{Rubric-as-reward vs.\ rubric-as-gate within \myname{}. Both variants use the same \rwdname{} dense reward and the same query-specific rubrics; they differ only in whether rubric judgments enter the gradient as a graded scalar reward or are used as hard rollout-group accept/reject constraints. Win rate (\%) vs.\ the base model.}
\label{tab:rubric_reward_vs_gate}
\small
\setlength{\tabcolsep}{8pt}
\renewcommand{\arraystretch}{1.15}
\begin{tabular}{lcc}
\toprule
\textbf{Method} & \textbf{\para{}} & \textbf{\rar{}} \\
\midrule
Rubric reward only (Table~\ref{tab:main_results}, row \#5) & 55.9 & 53.4 \\
\rwdname{} only (Table~\ref{tab:main_results}, row \#6) & 57.8 & 54.2 \\
\rwdname{} + Rubric reward & 54.6\textsuperscript{$\dagger$} & 52.1 \\
\rwdname{} + Rubric-gating (\myname{}) & \textbf{63.7} & \textbf{57.6} \\
\bottomrule
\end{tabular}
\vspace{2pt}
\\
\footnotesize
\textsuperscript{$\dagger$}Exhibits the no-revision mode collapse described in Appendix~\ref{app:no_revision}.
\end{table}

\section{Rubric Satisfaction}
\label{app:heldout_rubric}

\begin{figure}[h]
  \centering
  \includegraphics[width=0.45\textwidth]{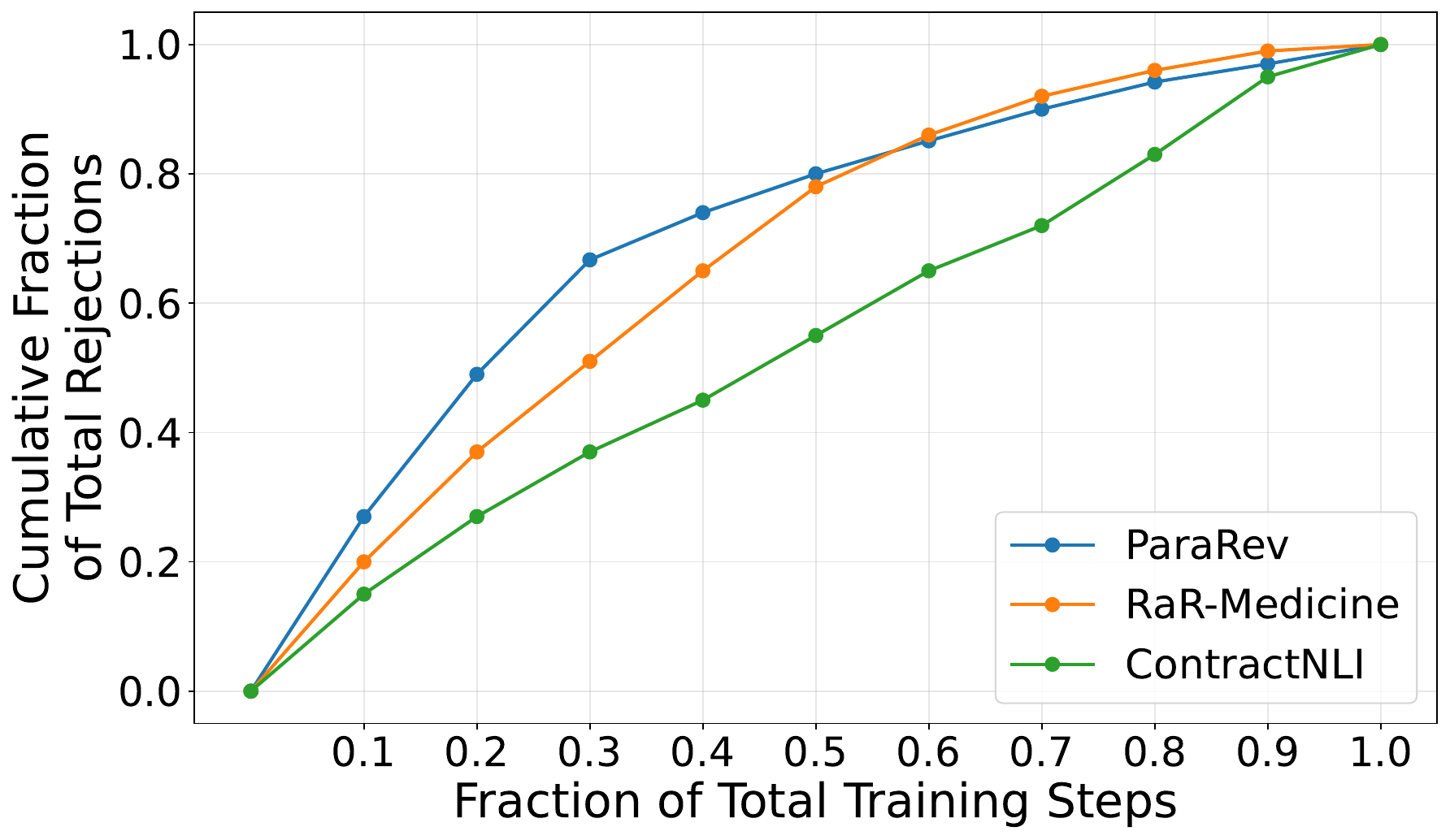}
  \caption{\textbf{Rubric rejection rate over training.} Cumulative fraction of total rejections. On \para{} and \rar{}, the curve flattens over training, indicating fewer rejections and improved rubric satisfaction. On \nda{}, the curve is closer to linear, suggesting limited improvement, consistent with downstream results.}
  \label{fig:rubric_convergence}
\end{figure}

In Section~\ref{sec:eval_rubric_gating} we monitor training-time constraint-satisfaction regret (Fig.~\ref{fig:rubric_convergence}), following practice in constrained RL~\citep{ray2019benchmarking}. As a generalization check, we additionally evaluate per-rubric satisfaction on \emph{held-out test queries}: for each test query we generate one response from the model and judge it against the same query-specific rubrics used during training, reporting the fraction of rubrics satisfied averaged over the test set. \myname{} substantially improves held-out rubric satisfaction over the base model on all three rubric-gated datasets (Table~\ref{tab:heldout_rubric}), indicating that rubric-gating drives generalizable constraint satisfaction rather than query-specific memorization.

\begin{table}[h]
\centering
\caption{Held-out rubric satisfaction (\% of query-specific rubrics satisfied on the test set).}
\label{tab:heldout_rubric}
\small
\setlength{\tabcolsep}{8pt}
\renewcommand{\arraystretch}{1.15}
\begin{tabular}{lcc}
\toprule
\textbf{Dataset} & \textbf{Base} & \textbf{\myname{}} \\
\midrule
\para{}     & 46.2\% & \textbf{81.4\%} \\
\rar{}      & 51.7\% & \textbf{77.6\%} \\
\nda{}      & 76.3\% & \textbf{87.2\%} \\
\bottomrule
\end{tabular}
\end{table}

\section{Variance-Based Filtering Improves Training Dynamics}
\label{app:stablize_reward}

\begin{figure}[t]
    \centering
    \includegraphics[width=0.4\linewidth]{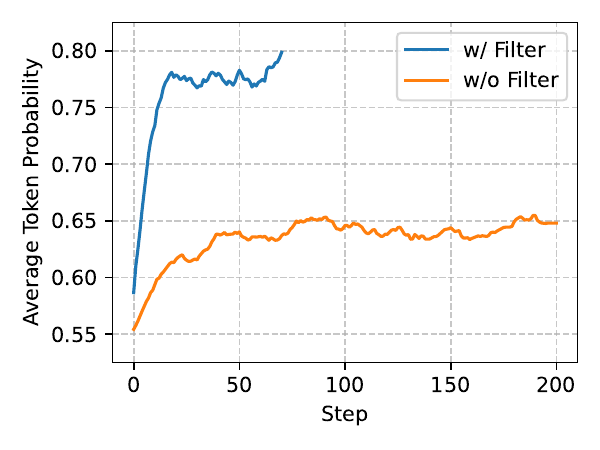}
    \caption{Average token probability during training with and without variance-based filtering. Filtering accelerates progress and yields earlier and more stable convergence.}
    \label{fig:avg_token_prob}
\end{figure}

Figure~\ref{fig:avg_token_prob} compares training with and without variance-based filtering. Filtering leads to faster improvement and a higher final average token probability, indicating earlier and more stable convergence.

\section{Comparison with RLPR's Variance-Based Filtering}
\label{app:rlpr_filter_comparison}

The concurrent RLPR work~\citep{yu2025rlpr} also performs prompt-level filtering by reward variance, but averages variance over \emph{all} reference tokens. Because uninformative tokens dominate long-form references (the same dilution problem that motivates \rwdname{}, Section~\ref{sec:aggregation}), all-token variance is itself diluted, so groups uninformative \emph{on the reasoning-reflective tokens} can still pass the filter. \myname{} instead restricts variance to reasoning-reflective tokens (top-$10\%$ by per-token cross-rollout standard deviation), better aligning the filter with the actual learning signal. Table~\ref{tab:rlpr_filter} compares the two filters on top of \rwdname{} (without rubric-gating), holding everything else fixed. Token-selective filtering improves over plain average-variance filtering on both datasets.

\begin{table}[h]
\centering
\caption{Variance-based filtering: token-selective (DRO) vs.\ all-token average (RLPR), applied on top of \rwdname{} without rubric-gating.}
\label{tab:rlpr_filter}
\small
\begin{tabular}{lcc}
\toprule
Filtering variant & \para{} (WR vs.\ Base) & \rar{} (WR vs.\ Base) \\
\midrule
\rwdname{} + no filtering (Table~\ref{tab:main_results}, row \#6) & 57.8 & 54.2 \\
\rwdname{} + RLPR plain average variance & 58.6 & 55.2 \\
\rwdname{} + DRO token-selective variance & \textbf{59.8} & \textbf{55.7} \\
\bottomrule
\end{tabular}
\end{table}